\documentclass[fleqn,12pt]{wlscirep}
\usepackage[utf8]{inputenc}
\usepackage[T1]{fontenc}
\usepackage{upgreek}
\usepackage{cite}
\usepackage{ragged2e}
\usepackage{titlesec}
\usepackage[colorlinks=true]{hyperref}
\usepackage{cleveref}
\usepackage{jabbrv}
\makeatletter
\InputIfFileExists{jabbrv-ltwa-en.ldf}{}{%
  \PackageError{jabbrv}{Missing jabbrv-ltwa-en.ldf}{}%
}
\makeatother
\newcounter{customfig}% define a new counter
\newcounter{mainfig}% define a new counter
\newcounter{suppfig}% define a new counter
\title{%
  {\fontsize{15.5}{12}\selectfont\centering
   SolarSeer: Ultrafast and accurate 24-hour solar irradiance forecasts outperforming numerical weather prediction across the USA\par}}
\author[1+, *]{\mbox{\small Mingliang Bai}}
\author[1, +]{\mbox{\small Zuliang Fang}}
\author[2, 3, 1, +]{\mbox{\small Shengyu Tao}}
\author[1]{\mbox{\small Siqi Xiang}}
\author[1]{\mbox{\small Jiang Bian}}
\author[4, 1]{\mbox{\small Yanfei Xiang}}
\author[1]{\mbox{\small Pengcheng Zhao}}
\author[1]{\mbox{\small Weixin Jin}}
\author[1]{\mbox{\small Jonathan A. Weyn}}
\author[1, *]{\mbox{\small Haiyu Dong}}
\author[1]{\mbox{\small Bin Zhang}}
\author[1]{\mbox{\small Hongyu Sun}}
\author[1, *]{\mbox{\small Kit Thambiratnam}}
\author[1, *]{\mbox{\small Qi Zhang}}
\author[4, 5, *]{\mbox{\small Hongbin Sun}}
\author[2, *]{\mbox{\small Xuan Zhang}}
\author[2, *]{\mbox{\small Qiuwei Wu}}

\affil[1]{Microsoft}
\affil[2]{Tsinghua Shenzhen International Graduate School, Shenzhen, China}
\affil[3]{University of California, Berkeley, USA}
\affil[4]{Tsinghua University, Beijing, China}
\affil[5]{Taiyuan University of Technology, Taiyuan, China}
\affil[*]{Corresponding.author, Email: mingliangbai@outlook.com, haiyu.dong@microsoft.com, 
kitth@microsoft.com, qizhang@microsoft.com, shb@tsinghua.edu.cn, xuanzhang@sz.tsinghua.edu.cn,
qiuwu@sz.tsinghua.edu.cn}
\affil[+]{These authors contributed equally. Mingliang Bai, Yanfei Xiang and Shengyu Tao finished this work during internship at Microsoft}

\begin{abstract}
\hspace*{1em}
\fontfamily{ptm}\selectfont % Set font to Times New Roman
Accurate 24-hour solar irradiance forecasting is essential for the safe and economic operation of solar photovoltaic systems. Traditional numerical weather prediction (NWP) models represent the state-of-the-art in forecasting performance but rely on computationally costly data assimilation and solving complicated partial differential equations (PDEs) that simulate atmospheric physics. Here, we introduce SolarSeer, an end-to-end large artificial intelligence (AI) model for solar irradiance forecasting across the Contiguous United States (CONUS). SolarSeer is designed to directly map the historical satellite observations to future forecasts, eliminating the computational overhead of data assimilation and PDEs solving. This efficiency allows SolarSeer to operate over 1,500 times faster than traditional NWP, generating 24-hour cloud cover and solar irradiance forecasts for the CONUS at 5-kilometer resolution in under 3 seconds. Compared with the state-of-the-art NWP in the CONUS, i.e., High-Resolution Rapid Refresh (HRRR), SolarSeer significantly reduces the root mean squared error of solar irradiance forecasting by 27.28\% in reanalysis data and 15.35\% across 1,800 stations. SolarSeer also effectively captures solar irradiance fluctuations and significantly enhances the first-order irradiance difference forecasting accuracy. SolarSeer’s ultrafast, accurate 24-hour solar irradiance forecasts provide strong support for the transition to sustainable, net-zero energy systems.

\end{abstract}

\begin{document}
\flushbottom
\maketitle
\thispagestyle{empty}

\section*{Main}
\addcontentsline{toc}{section}{Main}
\phantomsection
\label{Main}
\hspace*{1em}
% cite figures
% As shown in \hyperref[extend_fig1]{Extended Data Fig. 1}, this is a reference to the figure.
% As shown in \hyperref[extend_fig2]{Extended Data Fig. 2}, this is a reference to the figure.
% As shown in \hyperref[extend_fig3]{Extended Data \hyperref[fig4]{Fig. 4}}, this is a reference to the figure.
Global solar photovoltaic (PV) power generation has experienced a surge, increasing more than 40 times since 2010, as countries push towards carbon neutrality and mitigate climate change~\cite{IEA2023, zhang2025machine}.  In 2023, global grid-connected new solar PV capacity reached 447 GW, accounting for 78\% of the newly added global power generation capacity~\cite{solarpowereurope2024}. By the mid-century, solar energy is promising to become the world's dominant power source~\cite{nijsse2023momentum}.  However, the output of solar PV power is inherently tied to the solar irradiance at the ground surface, which has strong inherent fluctuations and uncertainties~\cite{wang2023inherent} due to irregular formation, motion, deformation, and dissipation of clouds~\cite{xia2024accurate}. This fluctuation brings significant challenges to the stability and safety of electricity grids, with abrupt changes in PV power potentially causing severe imbalances between supply and demand or even leading to sudden electricity shutdowns. Therefore, accurate forecasting of solar irradiance is critical to ensuring the safe, reliable, stable, and economic operation of electricity grids, especially in future high-penetration PV power systems~\cite{gandhi2024value}. 
%Global solar photovoltaic (PV) generation has increased more than 40 times since 2010 to push towards carbon neutrality and mitigate climate change~\cite{IEA2023}.  In 2023, the global grid-connected new solar PV capacity reached 447 GW, accounting for 78 \% of the newly added global power generation capacity~\cite{solarpowereurope2024}. By mid-century, solar energy is promising to become the dominant power source~\cite{nijsse2023momentum}. Solar PV power is directly determined by the solar irradiance reaching the ground surface, which exhibits significant fluctuation and intermittency due to irregular motion, formation, deformation, and dissipation of clouds ~\cite{xia2024accurate}, causing substantial imbalances between load demand and PV power generation and seriously threatening the safety and stability of the electricity grid. An abrupt ramp of PV power can even cause a sudden electricity shutdown. Accurate forecasting of solar irradiance is essential to address these challenges and ensure safe, reliable, stable, and economical operation of the electricity grid, especially in future high-penetration PV power systems ~\cite{gandhi2024value}. %

\begin{figure}[ht]
\centering
\includegraphics[width=\linewidth]{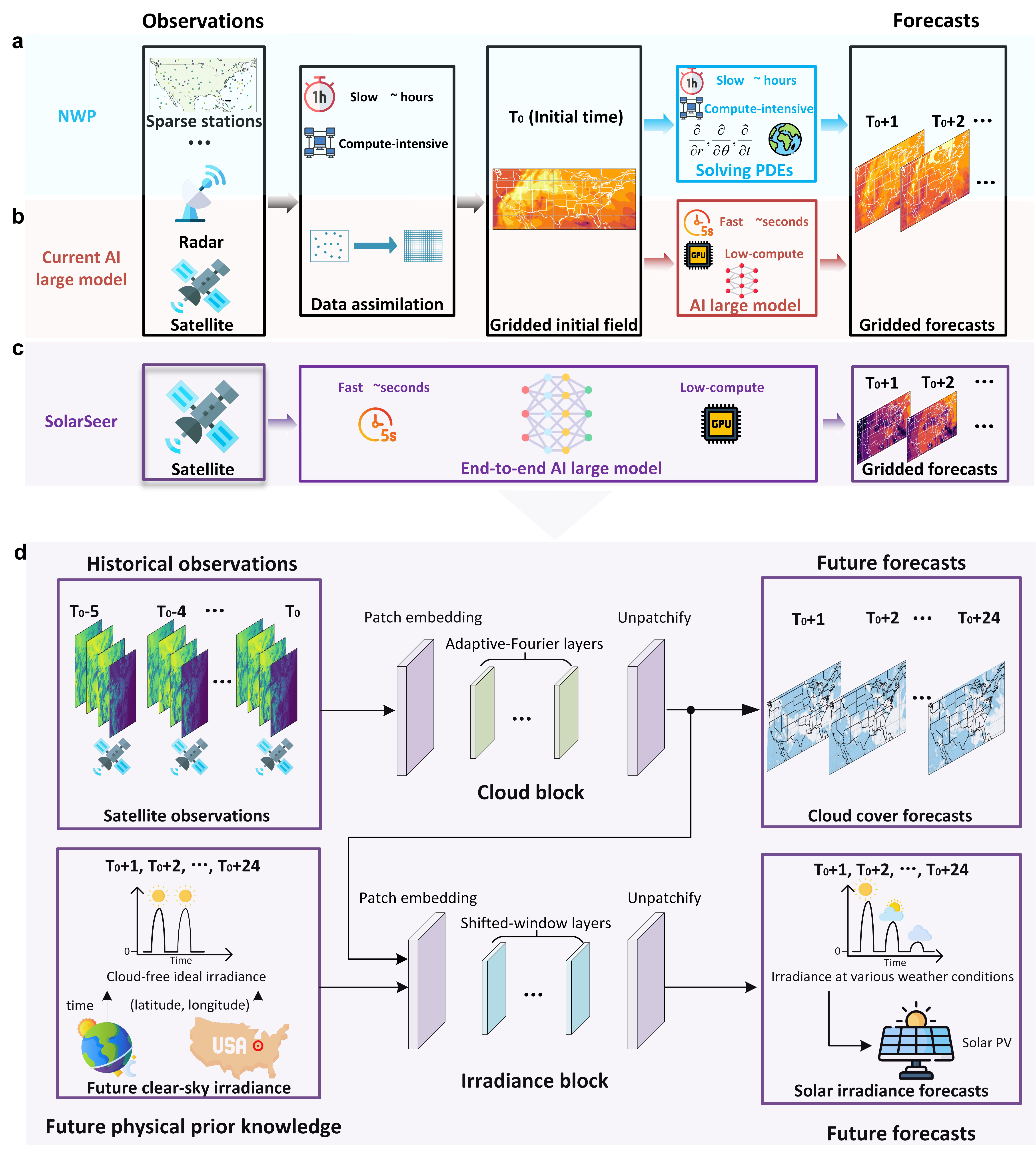}
\caption*{\justifying \textbf{Fig. \themainfig  \space \space SolarSeer is an end-to-end AI model for solar irradiance forecasting.} \textbf{(a-c)} Comparison of \textbf{(c)} SolarSeer with \textbf{(a)} numerical weather prediction (NWP) and \textbf{(b)} current AI large models. SolarSeer directly maps the satellite observations to gridded forecasts without the need for data assimilation and PDEs solving.  \textbf{(d)} Overall structure of SolarSeer. SolarSeer uses historical 6-hour satellite observations as inputs to produce 24-hour forecasts of total cloud cover and solar irradiance. }
\label{framework}
\stepcounter{mainfig}
\end{figure}

Accurate solar irradiance forecasting relies on the ability to predict cloud cover and other atmospheric phenomena.  Numerical weather prediction (NWP)~\cite{bauer2015quiet} is widely regarded as the state-of-the-art method for weather forecasting. As illustrated in \hyperref[framework]{Fig. 1} (a), NWP starts with data assimilation~\cite{zhou2024progress}, which transforms raw weather observations - collected from satellites, weather stations, radars, and other sources - into gridded initial fields. The initial fields are then passed through iterative methods to solve partial differential equations (PDEs) that represent atmospheric physics, ultimately producing gridded forecasts for key weather variables~\cite{pu2019numerical}. However, both data assimilation and PDE solving are computationally intensive, requiring significant resources and time to generate predictions. NWP usually runs on supercomputing clusters and needs one to several hours to generate forecasts. Artificial intelligence (AI) offers an alternative to NWP for solar irradiance forecasting by learning the temporal dependency among historical data directly, without relying on the computationally expensive steps of data assimilation and PDE solving. Although various AI-based methods for solar irradiance forecasting have been proposed~\cite{chen2024one,kumari2021deep,zhu2025novel}, most are designed for one or a few locations, offer reliable accuracy only for very short lead times (typically up to 4 hours), and fail to surpass NWP models in 24-hour forecasting performance~\cite{yang2022review}. This limitation restricts their applicability primarily to intraday electricity markets~\cite{srivastava2011electricity}, which require one to several hours' forecasts. However, most of the electricity trading—approximately 80\%—occurs in day-ahead markets~\cite{xia2024accurate}, which require reliable 24-hour forecasts. As such, existing AI-based methods are currently unable to meet the operational and economic requirements of modern electricity grids dominated by renewable energy sources, such as solar PV power.

Recent advances in large AI models have demonstrated their potential to surpass traditional NWP methods for specific weather variables—such as wind speed, temperature, pressure, and precipitation—at lead times of one to several days over global and regional domains~\cite{bi2023accurate,lam2023learning,price2025probabilistic, chen2023fuxi,bodnar2025foundation, espeholt2022deep}. These models (e.g., Pangu-Weather~\cite{bi2023accurate}, GraphCast~\cite{lam2023learning}, GenCast~\cite{price2025probabilistic}, Fuxi~\cite{chen2023fuxi}, Aurora~\cite{bodnar2025foundation}, and MetNet-2~\cite{espeholt2022deep}) utilize large-scale training datasets and combine spatial-temporal information to learn a direct mapping from gridded initial fields to future gridded weather forecasts (see \hyperref[framework]{Fig. 1} (b)). Despite these breakthroughs, no large AI model has yet been developed to forecast solar irradiance or cloud cover, let alone outperform NWP for 24-hour solar irradiance forecasts. Furthermore, these AI models continue to depend on the gridded initial fields produced by data assimilation, which introduces two major drawbacks. First, their reliance on data assimilation preserves the high computational cost associated with NWP~\cite{fisher2009data}, undermining the efficiency benefits of AI-based methods. Second, this dependence on preprocessed initial fields limits the ability of these models to operate as fully end-to-end forecasting systems, constraining their practical applicability for operational weather forecasting~\cite{schultz2021can}. Consequently, there is a persistent gap in the development of AI systems that can provide accurate, efficient, and standalone 24-hour solar irradiance forecasts suitable for day-ahead electricity markets.
%With large-scale training data and large-context spatial-temporal information fusion, recent breakthroughs in AI large models like Pangu-Weather~\cite{bi2023accurate}, GraphCast~\cite{lam2023learning}, Fuxi~\cite{chen2023fuxi}, Metnet-2~\cite{espeholt2022deep}, etc., which used AI to learn a mapping from gridded initial fields to gridded forecasts of future weather(see  \hyperref[framework]{Fig. 1} (b)), have achieved better accuracy than NWP for the forecasting of wind speed, temperature, pressure, humidity and precipitation. However, current AI large models have not provided any forecast of cloud cover and solar irradiance. Thus, there are still no AI large models outperforming NWP in 24-hour solar energy forecasts or serving day-ahead electricity markets. Meanwhile, current AI large models still rely on the gridded initial fields generated by data assimilation (see \hyperref[framework]{Fig. 1} (b)). This limitation results in two main issues. First, current large AI models are unable to avoid the substantial computational costs associated with data assimilation. Second, these models still reply on the initial conditions as inputs, hindering their use in operational weather forecasting~\cite{schultz2021can}. Consequently, there remains a gap for an accurate, fully AI-based model capable of serving operational 24-hour solar energy forecasting and day-ahead electricity markets.

Here we introduce SolarSeer, the first end-to-end AI-based model, to our knowledge, that outperforms state-of-the-art NWP for 24-hour-ahead solar irradiance forecasting at kilometer-scale resolution. SolarSeer (see \hyperref[framework]{Fig. 1}(c)) directly learns a mapping from historical satellite observations to forecasts of future cloud cover and solar irradiance, completely bypassing both data assimilation and PDE solving during the inference phase. As a result, SolarSeer operates more than 1,500 times faster than NWP, generating 24-hour forecasts at 5-kilometer resolution in less than 3 seconds.  Compared with the state-of-the-art NWP in the Contiguous United States (CONUS), i.e., High-Resolution Rapid Refresh
HRRR)~\cite{dowell2022high, james2022high}, SolarSeer significantly reduces the root mean squared error (RMSE) of solar irradiance forecasting by 27.28\% in reanalysis data and by 15.35\% across 1,800 stations in CONUS. Our model can be applied to the operational forecast of solar PV power generation, greatly reduce the economic costs of inaccurate forecasting, and significantly improve the competitiveness of solar PV in day-ahead electricity markets.
%Here we introduce SolarSeer, an end-to-end AI model that outperforms the start-of-the-art NWP for 24-hour-ahead kilometer-scale solar energy forecasting. SolarSeer (see \hyperref[framework]{Fig. 1}(c)) uses AI to learn a mapping from the historical satellite observations to the future cloud cover and solar irradiance, avoids the huge computational costs of both data assimilation and PDEs solving, and thus runs more than 1,500 times faster than NWP. In under 3 seconds, SolarSeer can provide 24-hour forecasting of total cloud cover and solar irradiance at 5-kilometer spatial resolution, significantly outperforming state-of-the-art NWP in the contiguous United States, i.e., High-Resolution Rapid Refresh (HRRR), in reanalysis data. Our model can be applied to the operational forecast of solar PV power generation and significantly improve the competitiveness of solar PV in electricity markets.

\section*{Results}
\addcontentsline{toc}{section}{Results}
\phantomsection
\label{Results}

\subsection*{SolarSeer overview}
\phantomsection
\addcontentsline{toc}{subsection}{SolarSeer overview}
\label{SolarSeer_overview}
\hspace*{1em}
We present SolarSeer, an end-to-end AI-based framework outperforming NWP for solar energy forecasting (see \hyperref[framework]{Fig. 1}(c)). SolarSeer uses historical 6-hour satellite observations as input to generate future 24-hour cloud cover and solar irradiance forecasts with a high spatial resolution of 0.05° (approximately 5 km) across the contiguous United States. Specifically, SolarSeer contains the cloud block and the irradiance block as follows (see \hyperref[framework]{Fig. 1} (d)):

The cloud block contains multiple Adaptive Fourier Neural Operator (AFNO) Transformer layers and maps the historical 6-hour satellite images to the future 24-hour cloud cover forecasts. The irradiance block comprises multiple Swin Transformer layers, which map the cloud block output and the future 24-hour clear-sky solar irradiance to the future 24-hour solar irradiance forecasts (see \hyperref[network]{SolarSeer network framework}). Note that clear-sky solar irradiance, estimated by the Ineichen-Perez model, is a simple function of longitude, latitude, and time due to the rotation and revolution of the Earth (see \hyperref[clear-sky]{Clear-sky solar irradiance model}).  Clear-sky solar irradiance characterizes the maximum possible solar irradiance reaching the ground under cloud-free situations and serves as the future physical prior knowledge for solar irradiance forecasting. 

SolarSeer is trained using over five years’ data and 16 AMD MI200 GPUs for about one week. In under 3 seconds, SolarSeer can give 24-hour forecasting of cloud cover and solar irradiance, which is more than 1,500 times faster than the state-of-the-art NWP in the Contiguous United States (CONUS), i.e., High-Resolution Rapid Refresh (HRRR) \cite{dowell2022high, james2022high} of the National Oceanic and Atmospheric Administration (NOAA).  For more details about the data and network structure, see \hyperref[data]{Data} and \hyperref[network]{SolarSeer network framework}.

\subsection*{Cloud cover forecast results}
\phantomsection
\addcontentsline{toc}{subsection}{Cloud cover forecast results}
\label{Cloud_cover_forecast_results}
\hspace*{1em}
We compared the cloud cover forecasting of SolarSeer with HRRR-NWP ~\cite{dowell2022high, james2022high}. HRRR-NWP generates weather forecasts hourly, providing 48-hour forecasts at the initial times UTC 00:00, 06:00, 12:00, and 18:00, and 18-hour forecasts at other initial times (e.g., UTC 01:00, UTC 07:00, etc.) throughout the day. In contrast, SolarSeer generates forecasts hourly, providing 24-hour forecasts at all initial times of the day. Consequently, we evaluated the 24-hour forecasts of both SolarSeer and HRRR-NWP at the initial times UTC 00:00, 06:00, 12:00, and 18:00, while evaluating the 24-hour forecasts of SolarSeer and the 18-hour forecasts of HRRR-NWP at other initial times throughout the day. The Mean Absolute Error (MAE) and the Root Mean Square Error (RMSE) are used to evaluate the forecast results (see \hyperref[metrics]{Evaluation metrics}), where the Real-Time Mesoscale Analysis (RTMA) dataset ~\cite{de2011real} is used as the ground truth of the total cloud cover. Smaller MAE and RMSE represent better forecast performance. 

\hyperref[fig2]{Fig. 2} (a) and \hyperref[extend_fig1]{Extended Data Fig. 1} (a) present the variation of RMSE and MAE, respectively, in total cloud cover forecasts as a function of lead time across all initial times of the day (UTC 00:00 to UTC 23:00). SolarSeer consistently outperforms the HRRR-NWP model across all lead times (1 to 24 hours) and initial times (UTC 00:00 to UTC 23:00). The spatial distribution of the total cloud cover forecasting errors for SolarSeer and HRRR-NWP across CONUS is presented in \hyperref[fig3]{Fig. 3} (a) (RMSE metric) and \hyperref[extend_fig2]{Extended Data Fig. 2} (a) (MAE metric). Compared with HRRR-NWP, SolarSeer achieves substantially smaller RMSE in more than 99.99\% regions of CONUS (see \hyperref[fig3]{Fig. 3} (c)) and substantially smaller MAE in 99.98\% regions of CONUS (see \hyperref[extend_fig2]{Extended Data Fig. 2}(c)). SolarSeer reduces RMSE by approximately 20\% compared to HRRR-NWP across most regions of CONUS, with reductions exceeding 30\% in certain areas along the western coast. (See \hyperref[fig3]{Fig. 3} (c)). Compared with HRRR-NWP, SolarSeer reduces the average RMSE by 29.13\% and the average MAE by 25.24\% for the forecast of 24-hour cloud cover in CONUS.

An example of 24-hour total cloud cover forecasts is presented in \hyperref[extend_fig3]{Extended Data Fig. 3} (a). The comparison includes total cloud cover forecasts over CONUS at lead times of 1, 3, 6, 9, 12, 18, and 24 hours from SolarSeer and HRRR-NWP, alongside the ground truth provided by RTMA reanalysis data, all initialized at UTC 06:00 on January 20, 2023. Across all forecast lead times, SolarSeer, HRRR-NWP, and the ground truth exhibit broadly similar spatial distributions of total cloud cover, with SolarSeer's forecasts demonstrating a closer alignment with the ground truth compared to those from HRRR-NWP. HRRR-NWP tends to predict total cloud cover as nearly completely cloudy (100\%) or fully sunny (0\%), while there are some intermediate values (e.g., 53\%) in ground truth. SolarSeer effectively predicts these intermediate values. Furthermore, HRRR-NWP fails to predict some regions with small values of total cloud cover (marked by red circles in \hyperref[extend_fig3]{Extended Data Fig. 3} (a)), while SolarSeer predicts them successfully. This case study demonstrates that SolarSeer provides better cloud cover forecasting than HRRR-NWP.

%An example of 24-hour total cloud cover forecast is presented in \hyperref[fig4]{Fig.4} (a). It compares total cloud cover forecasts over CONUS for 1, 3, 6, 9, 12, 18, and 24 hours ahead from SolarSeer, HRRR-NWP, and the ground truth (i.e., RTMA reanalysis data) for the initial time of UTC 2023-01-20 06:00:00. SolarSeer, HRRR-NWP, and ground truth exhibit similar spatial distributions across all forecast lead times, with SolarSeer aligning more closely with the ground truth than HRRR-NWP. HRRR-NWP tends to predict total cloud cover as either nearly fully cloudy (100 \%) or fully sunny (0 \%), while there are some intermediate values (e.g.， 53 \%) in the ground truth. SolarSeer effectively predicts these intermediate values. Additionally, HRRR-NWP fails to predict some regions with small values of total cloud cover (marked by red circles in \hyperref[fig4]{Fig.4} (a)), while SolarSeer predicts them successfully. This case study demonstrates that SolarSeer provides better cloud cover forecasting than HRRR-NWP. james2017unified

\subsection*{Solar irradiance forecast results}
\phantomsection
\addcontentsline{toc}{subsection}{Solar irradiance forecast results}
\label{Solar_irradiance_forecast_results}
\hspace*{1em}
We compared SolarSeer not only with HRRR-NWP but also with clear-sky solar irradiance, where HRRR-NWP is the start-of-the-art solar irradiance forecasting~\cite{marquis2015public, james2017unified, zhang2022solar} in CONUS and clear-sky measures the maximum possible solar irradiance reaching the ground under cloud-free situations(see \hyperref[clear-sky]{Clear-sky solar irradiance model}). The fifth generation of ECMWF Reanalysis Data (ERA5)~\cite{hersbach2020era5} is used as the ground truth of solar irradiance. 

\begin{figure}[ht]
\centering
\includegraphics[width=\linewidth]{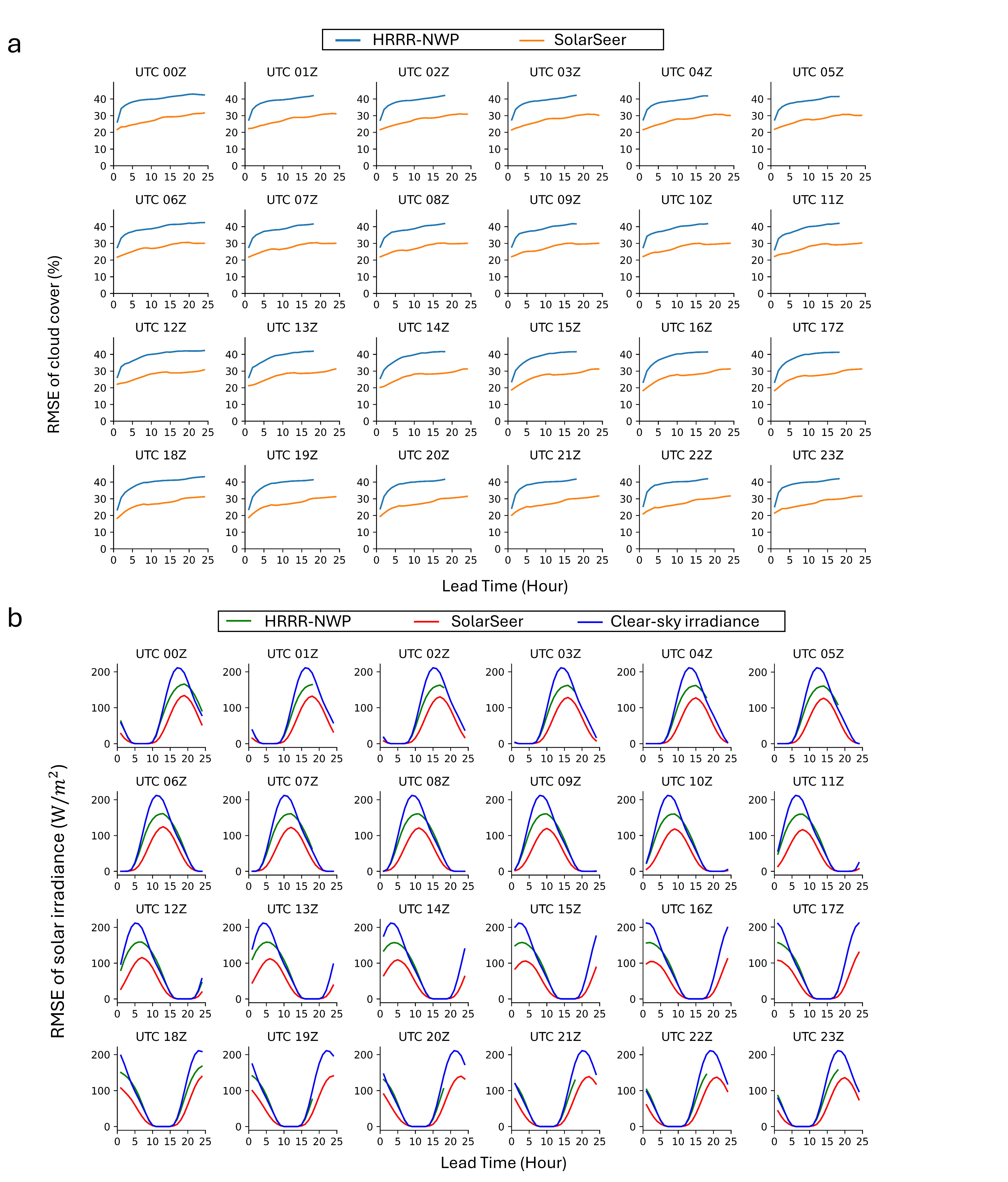}
\caption*{\justifying \textbf{Fig. \themainfig  \space \space SolarSeer demonstrates smaller 24-hour forecasting RMSE than HRRR-NWP across all initial times.} \textbf{(a)} RMSE of cloud cover forecasting. \textbf{(b)} RMSE of solar irradiance forecasting. HRRR-NWP provides 48-hour forecasts at the initial time of UTC 00:00, 06:00, 12:00, and 18:00 (designated as UTC 00Z, UTC 06Z, UTC 12Z, and UTC 18Z in the figure title), and 18-hour forecasts at other times. SolarSeer provides 24-hour forecasts for all initial times. This figure only demonstrates the future 24-hour forecasts.}
\label{fig2}
\stepcounter{mainfig}
\end{figure}

\begin{figure}[ht]
\centering
\includegraphics[width=\linewidth]{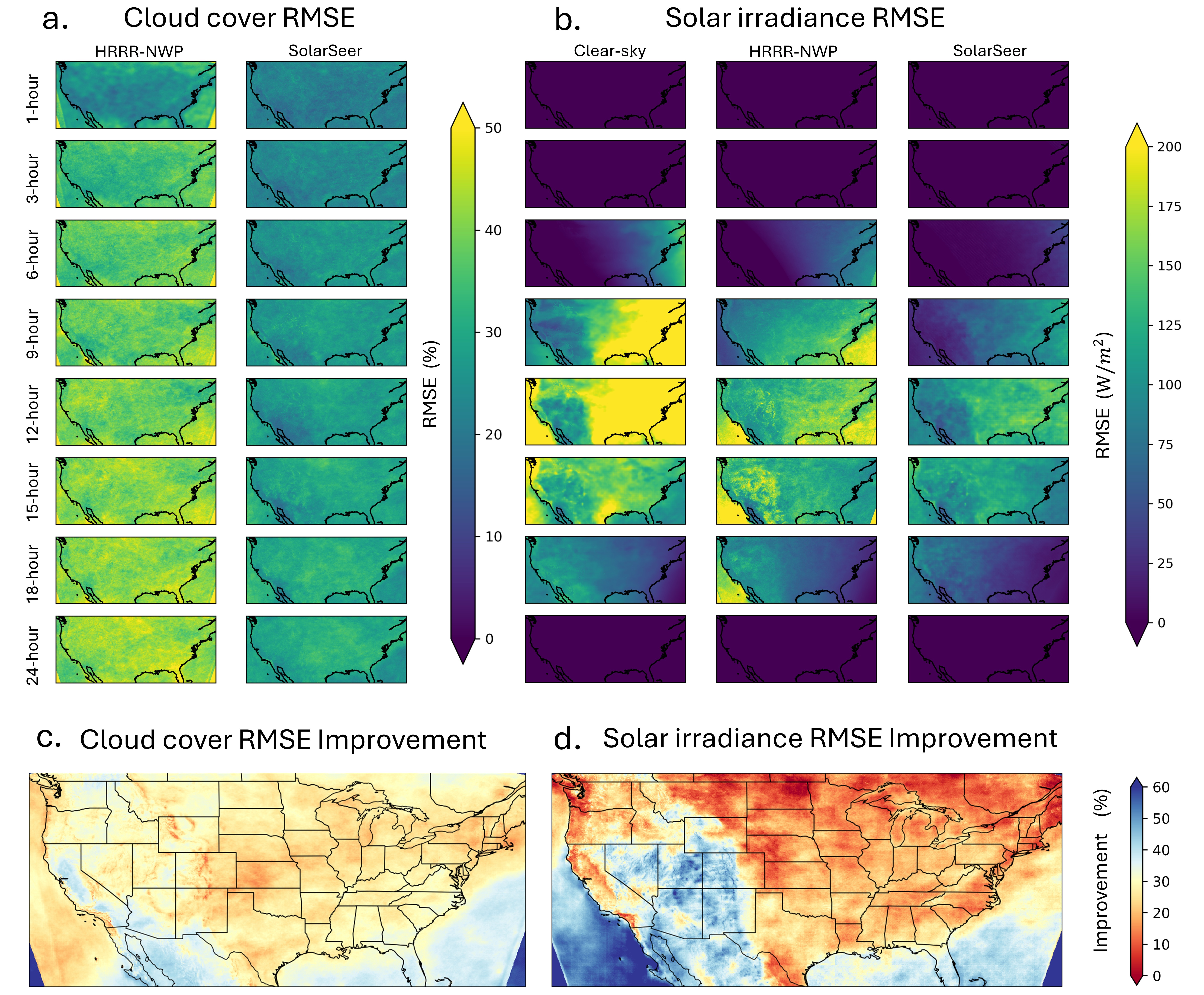}
\caption*{\justifying \textbf{Fig. \themainfig  \space \space SolarSeer demonstrates smaller forecasting RMSE than HRRR-NWP across the Contiguous United States.} \textbf{(a)} Spatial distribution of RMSE errors for cloud cover forecasting. \textbf{(b)} Spatial distribution of RMSE errors for solar irradiance forecasting. \textbf{(c)} Mean RMSE improvement of SolarSeer over HRRR-NWP for 24-hour cloud cover forecasting. \textbf{(d)} Mean RMSE improvement of SolarSeer over HRRR-NWP for 24-hour solar irradiance forecasting. In \textbf{(a)} and \textbf{(b)}, darker colors indicate smaller forecast RMSE errors, while lighter colors denote larger RMSE errors. The spatial error maps reveal that SolarSeer has consistently smaller forecast RMSE errors than HRRR-NWP across 1-hour, 3-hour, 6-hour, 9-hour, 12-hour, 15-hour, 18-hour, and 24-hour forecasts of cloud cover and solar irradiance. All the initial time is UTC 06:00:00 because it approximately corresponds to local time 00:00:00.}
\label{fig3}
\stepcounter{mainfig}
\end{figure}

The variation of RMSE and MAE versus the lead time for solar irradiance forecasts is presented in \hyperref[fig2]{Fig. 2} (b) and \hyperref[extend_fig1]{Extended Data Fig. 1} (b). Both SolarSeer (red lines in \hyperref[fig2]{Fig. 2}(b) and \hyperref[extend_fig1]{Extended Data Fig. 1}(b)) and HRRR-NWP (green lines in \hyperref[fig2]{Fig. 2}(b) and \hyperref[extend_fig1]{Extended Data Fig. 1}(b)) significantly outperform clear-sky solar irradiance (blue lines in \hyperref[fig3]{Fig. 2}(b) and \hyperref[extend_fig1]{Extended Data Fig. 1}(b)). This shows that they can both correct the cloud-free solar irradiance well because of the modeling of cloud cover. SolarSeer significantly outperforms HRRR-NWP in all lead times from 1 to 24 hours and at all initial times from UTC 00:00 to UTC 23:00. SolarSeer significantly outperforms HRRR-NWP in cloud cover forecasting (see \hyperref[fig2]{Fig. 2}(a) and \hyperref[extend_fig1]{Extended Data Fig. 1}(a)), and thus it can achieve significantly lower RMSE and MAE than HRRR-NWP in solar irradiance forecasting. 

The spatial distribution of RMSE and MAE in CONUS is shown in \hyperref[fig3]{Fig. 3} (b) and \hyperref[extend_fig2]{Extended Data Fig. 2}(b), respectively. Compared to HRRR-NWP, SolarSeer reduces RMSE in 99.87\% regions of CONUS (see \hyperref[fig3]{Fig. 3} (d)) and reduces MAE in 99.99\% regions of CONUS (see \hyperref[extend_fig2]{Extended Data Fig. 2} (d)). In most regions, SolarSeer lowers RMSE by about 10\%-20\%, with reductions of approximately 40\% in the western United States and around 30\% in southern regions such as the Florida Peninsula (see \hyperref[fig3]{Fig. 3} (d)). Compared with HRRR-NWP, SolarSeer reduces the average RMSE by 27.28\% and the average MAE by 36.08\% for the forecast of solar irradiance at 24 hours in CONUS. A comparison of \hyperref[fig3]{Fig. 3} (c) and (d) shows that SolarSeer achieves the greatest improvements in the western and southern regions for both cloud cover and solar irradiance forecasting, indicating that the improved solar irradiance forecasting by SolarSeer can be partially attributed to its improved cloud cover forecasting. 

Similar to cloud cover analysis, an example of solar irradiance forecasting is presented in \hyperref[extend_fig3]{Extended Data Fig. 3} (b), with a initial time of UTC 2023-01-20 06:00:00. Both SolarSeer and HRRR-NWP predictions are similar to the ground truth (ERA5 reanalysis data~\cite{hersbach2020era5}). Additionally, SolarSeer and HRRR-NWP are both more similar to ground truth than clear-sky irradiance because clear-sky irradiance does not consider any influence of clouds. A comparison of \hyperref[extend_fig3]{Extended Data Fig. 3} (a) and (b) shows that regions with high cloud cover tend to have low solar irradiance, highlighting the critical importance of accurate cloud cover forecasting for solar irradiance predictions.

\subsection*{Verification of solar irradiance forecast on weather stations}
\begin{figure}[ht]
\centering
\includegraphics[width=\linewidth]{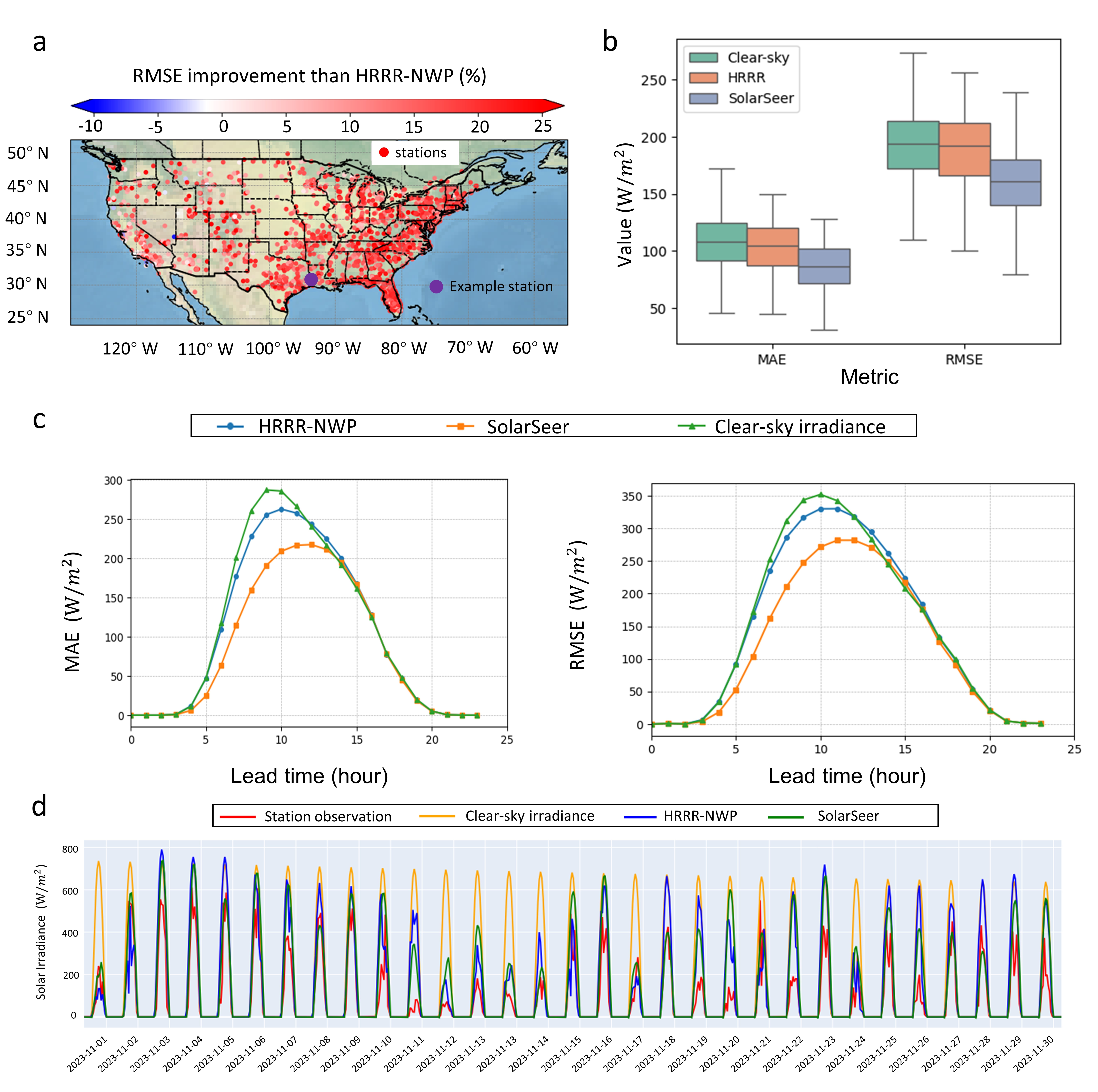}
\caption*{\textbf{Fig. \themainfig  \space \space SolarSeer outperforms HRRR-NWP in the solar irradiance forecasting at most weather stations across Contiguous United States.} \textbf{(a)} Spatial distribution of 1,800 weather stations across Contiguous United States. In 98.78\% stations, SolarSeer obtains smaller RMSE than HRRR-NWP. \textbf{(b)} Boxplot of solar irradiance forecasting errors for clear-sky irradiance, HRRR-NWP and SolarSeer. SolarSeer shows the smallest median MAE and RMSE of 1, 800 weather stations.  \textbf{(c)} MAE and RMSE comparison of clear-sky irradiance, HRRR-NWP and SolarSeer. SolarSeer achieves the lowest MAE and RMSE. \textbf{(d)} Visualization of solar irradiance forecasting for the League City weather station (marked in purple in \textbf{(a)}) in November 2023. SolarSeer (green lines) is much closer to the station observations (red lines) than HRRR-NWP (blue lines). All the initial time is UTC 06:00:00 in this figure because UTC 06:00:00 approximately corresponds to local time 00:00:00.}
\label{fig4}
\stepcounter{mainfig}
\end{figure}

\phantomsection
\addcontentsline{toc}{subsection}{Verification of solar irradiance forecast on weather stations}
\label{station_verification}
\hspace*{1em}
Although we have verified solar irradiance forecasting using ERA5 reanalysis data~\cite{hersbach2020era5}, observation data from weather stations generally provides a more accurate representation of actual solar irradiance than reanalysis data. Therefore, we also verified solar irradiance forecasts using solar irradiance observation data from 1,800 weather stations provided by Synoptic (see \hyperref[data]{Data}). The spatial distribution of the 1,800 weather stations across CONUS is shown in \hyperref[fig4]{Fig. 4} (a).

SolarSeer is trained exclusively on reanalysis data, without using any station observations during training. Although ERA5 reanalysis and in situ measurements exhibit systematic differences in solar irradiance patterns, SolarSeer nevertheless delivers superior forecast accuracy at weather stations compared to HRRR‐NWP. Specifically, SolarSeer achieves a lower RMSE than HRRR-NWP at 1,778 stations (98.78\% of the total), and its median MAE and RMSE are significantly lower than those of both HRRR-NWP and clear-sky irradiance(see \hyperref[fig4]{Fig. 4} (b)). For 24-hour forecasts, SolarSeer also outperforms both benchmarks in RMSE and MAE (see \hyperref[fig4]{Fig. 4} (c)). Unlike the reanalysis data results, where HRRR-NWP significantly outperforms clear-sky solar irradiance, HRRR-NWP has a slightly lower  MAE and RMSE than clear-sky irradiance at the weather stations (see \hyperref[fig4]{Fig. 4} (b) and (c)). This discrepancy may be due to differences between the solar irradiance patterns in ERA5 reanalysis data and station observations. Nonetheless, across both reanalysis and station data, SolarSeer consistently yields the lowest errors. Using station observations as ground truth at the initial time of UTC 06:00, SolarSeer reduces average MAE by 16.86\% and average RMSE by 15.35\% relative to HRRR-NWP and maintains significant improvements at all other initial times throughout the day. An example of solar irradiance forecasting for the weather station in League City is shown in \hyperref[fig4]{Fig. 4} (d), where SolarSeer matches station observations more closely than HRRR-NWP and clear-sky irradiance on most days of November 2023. 

The evaluation of 24-hour solar irradiance forecasting at 1,800 weather stations across CONUS demonstrates the superior performance of SolarSeer and its potential for operational solar energy forecasting.

\subsection*{Interpretation}
\phantomsection
\addcontentsline{toc}{subsection}{Interpretation}
\label{Interpretation}
\hspace*{1em}
In this section, we explain why SolarSeer can accurately predict solar irradiance. SolarSeer first uses the cloud block to map historical satellite observations to future cloud cover forecasts. It then employs the irradiance block to map future cloud cover and clear-sky irradiance to future solar irradiance.

Clear-sky irradiance represents the maximal possible solar irradiance under cloud-free conditions. It is primarily determined by the relative movement between the Sun and the Earth (See \hyperref[clear-sky]{Clear-sky solar irradiance model}). The absolute difference between clear-sky irradiance and actual solar irradiance represents the uncertain component of solar irradiance. This uncertainty arises mainly from the effects of clouds absorbing, reflecting, or scattering solar radiation. %The absolute difference between clear-sky irradiance and actual solar irradiance represents the uncertain component of solar irradiance, which is mainly influenced by clouds.%
As shown in \hyperref[extend_fig6]{Extended Data Fig. 6}  (a), the pattern of the uncertain component of solar irradiance is similar to the total cloud cover during the day. As shown in \hyperref[extend_fig6]{Extended Data Fig. 6} (b), there is a significant positive correlation between the uncertain component of solar irradiance and total cloud cover. In 85.15\% of the regions across CONUS, the correlation coefficient between the uncertain component of solar irradiance and the total cloud cover during the day is greater than 0.5. This indicates that better cloud forecasting can lead to more accurate irradiance forecasting. SolarSeer forecasts cloud cover more accurately than HRRR-NWP (\hyperref[fig3]{Fig. 3 (a) and (c)}), enabling it to produce more precise solar irradiance forecasts.

\subsection*{Benefits for future high-penetration photovoltaic power systems}
\phantomsection
\addcontentsline{toc}{subsection}{Benefits for future high-penetration photovoltaic power systems}
\label{Benefits}
\hspace*{1em}
Due to the inherent fluctuation of solar PV power generation, accurate forecasting of solar irradiance fluctuations is essential for effectively allocating flexibility resources in the power grid~\cite{kamadinata2020solar}. We use the first-order difference $Diff$ to characterize the fluctuation of solar irradiance and cloud cover and use MAE and RMSE to evaluate the first-order difference forecasting results (see \hyperref[metrics]{Evaluation metrics}). 

\begin{figure}[ht]
\centering
\includegraphics[width=\linewidth]{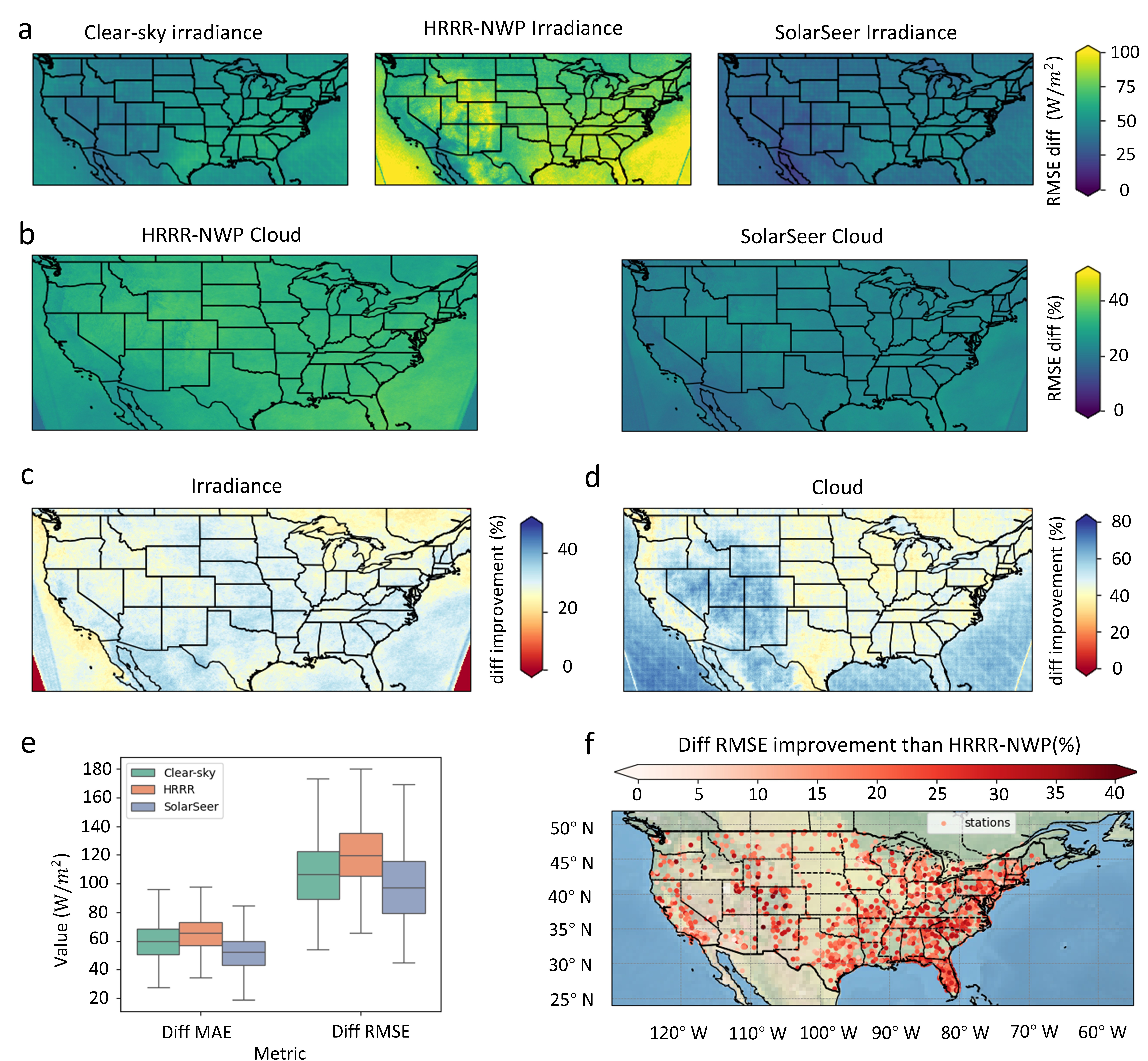}
\caption*{\justifying \textbf{Fig. \themainfig \space \space} SolarSeer demonstrates smaller first-order difference forecasting RMSE than HRRR-NWP. (a) RMSE of the solar irradiance first-order difference forecasting in reanalysis data. (b) RMSE of the cloud cover first-order difference forecasting in reanalysis data. (c) RMSE improvement of the solar irradiance first-order difference forecasting in reanalysis data. (d) RMSE improvement of the cloud cover first-order difference forecasting in reanalysis data. (e) Boxplot for MAE and RMSE of solar irradiance first-order difference forecasting in 1,800 weather stations. (f) RMSE improvement of solar irradiance first-order difference forecasting in 1,800 weather stations. All the initial time is UTC 06:00:00 because it approximately corresponds to local time 00:00:00.}
\label{fig5}
\stepcounter{mainfig}
\end{figure}

\begin{figure}[ht]
\centering
\includegraphics[width=\linewidth]{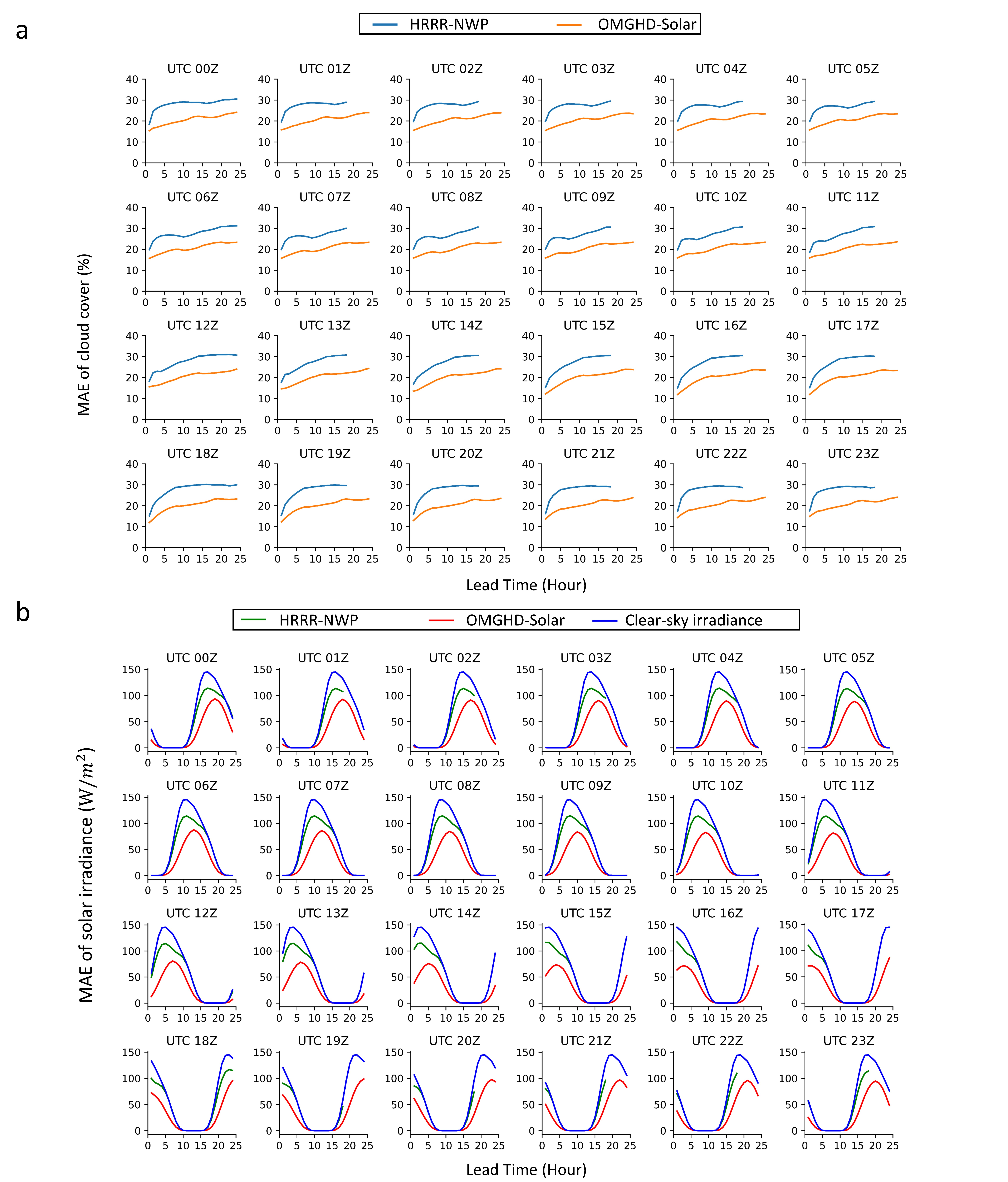}
\caption*{\justifying \textbf{Extended Data Fig. \thecustomfig \space \space} SolarSeer demonstrates smaller 24-hour forecasting MAE than HRRR-NWP across all initial times. (a) MAE of cloud cover forecasting. (b) MAE of solar irradiance forecasting. HRRR-NWP provides 48-hour forecasts at the initial time of UTC 00:00, 06:00, 12:00 and 18:00(designated as UTC 00Z, UTC 06Z, UTC 12Z and UTC 18Z in the figure title), and 18-hour forecasts at other times. SolarSeer provides 24-hour forecasts for all initial times. This figure only demonstrates the future 24-hour forecasts.}
\label{extend_fig1}
\stepcounter{customfig}
\end{figure}

\begin{figure}[ht]
\centering
\includegraphics[width=\linewidth]{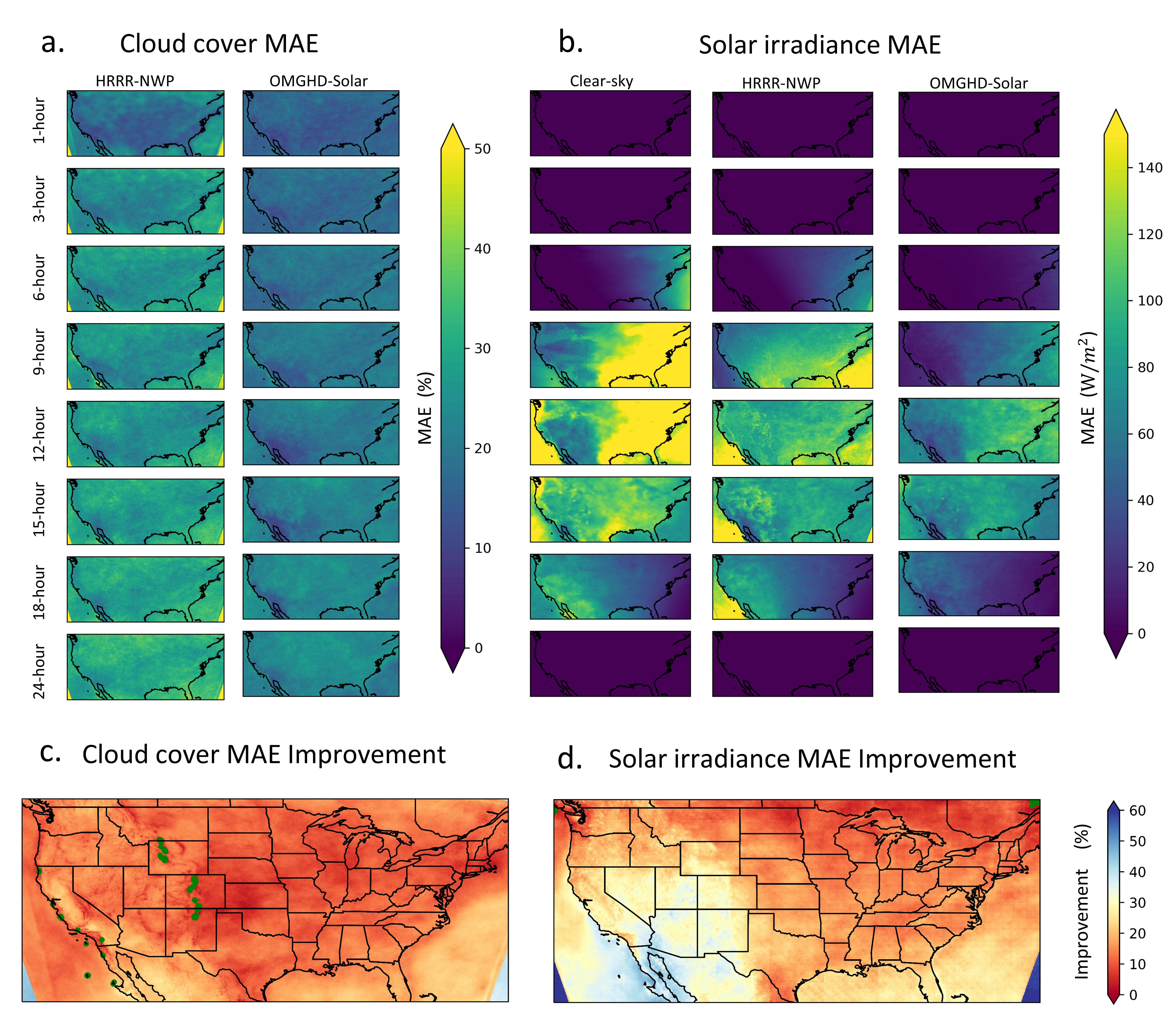}
\caption*{\justifying \textbf{Extended Data Fig. \thecustomfig \space \space} SolarSeer demonstrates smaller forecasting MAE than HRRR-NWP across the Contiguous United States. (a) Spatial distribution of MAE errors for cloud cover forecasting. (b) Spatial distribution of MAE errors for solar irradiance forecasting. (c) Mean MAE improvement of SolarSeer over HRRR-NWP for 24-hour cloud cover forecasting. (d) Mean MAE improvement of SolarSeer over HRRR-NWP for 24-hour solar irradiance forecasting. In (a) and (b), darker colors indicate smaller forecast MAE errors, while lighter colors denote larger MAE errors. The spatial error maps reveal that SolarSeer consistently has smaller forecast MAE errors than HRRR-NWP across 1-hour, 3-hour, 6-hour, 9-hour, 12-hour, 15-hour, 18-hour, and 24-hour forecasts of cloud cover and solar irradiance. All the initial time is UTC 06:00:00 because it approximately corresponds to local time 00:00:00.}
\label{extend_fig2}
\stepcounter{customfig}
\end{figure}

\begin{figure}[ht]
\centering
\includegraphics[width=\linewidth]{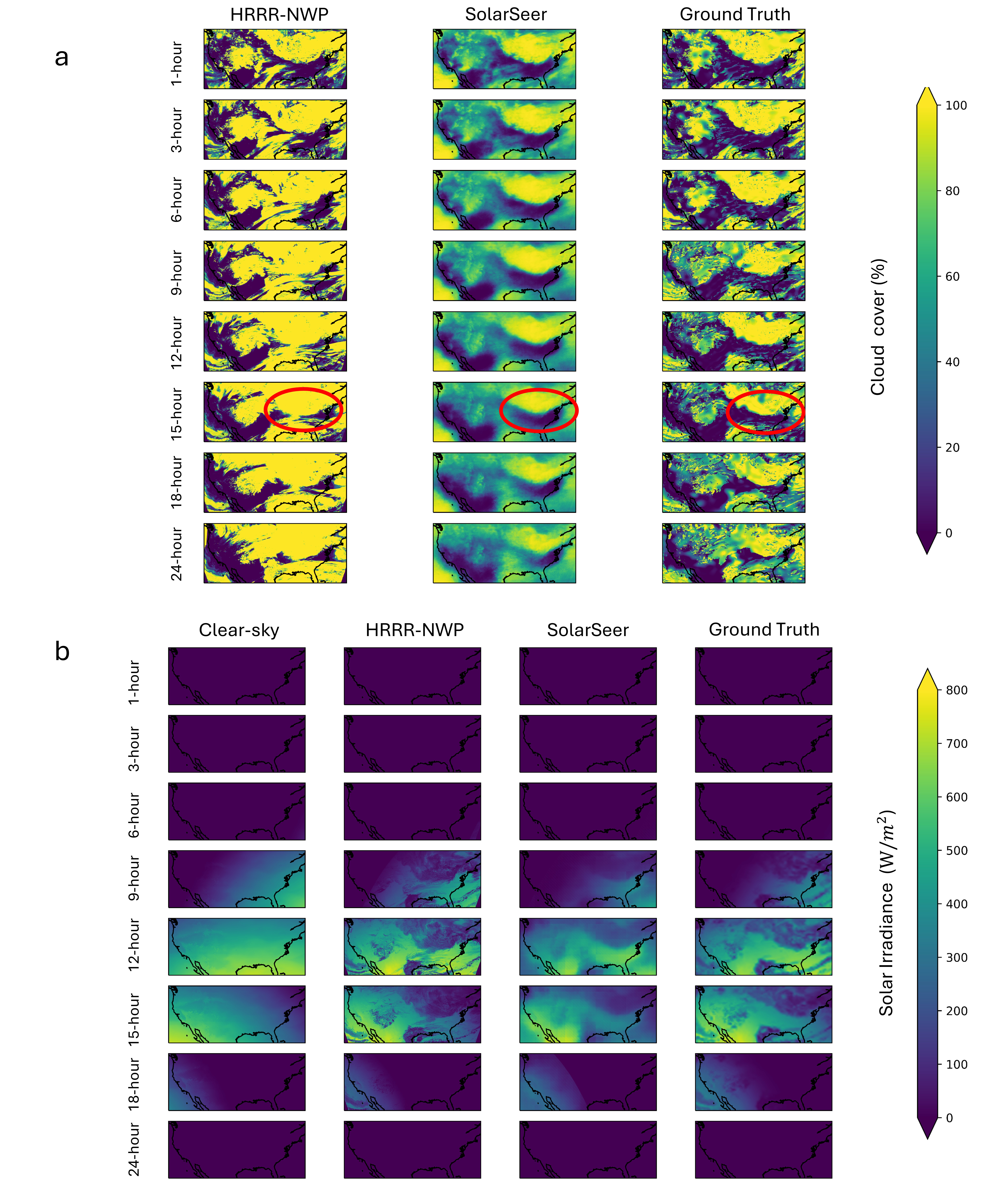}
\caption*{\justifying \textbf{Extended Data Fig. \thecustomfig \space \space} Visualization of forecasting results across the Contiguous United States. \textbf{(a)} Visualization of cloud cover forecasting. \textbf{(b)} Visualization of solar irradiance forecasting. The pattern of SolarSeer 1-hour, 3-hour, 6-hour, 9-hour, 12-hour, 15-hour, 18-hour, and 24-hour forecasting is similar to the ground truth. For all cases, the initial time is UTC 2023-01-20 06:00:00.}
\label{extend_fig3}
\stepcounter{customfig}
\end{figure}

\begin{figure}[ht]
\centering
\includegraphics[width=\linewidth]{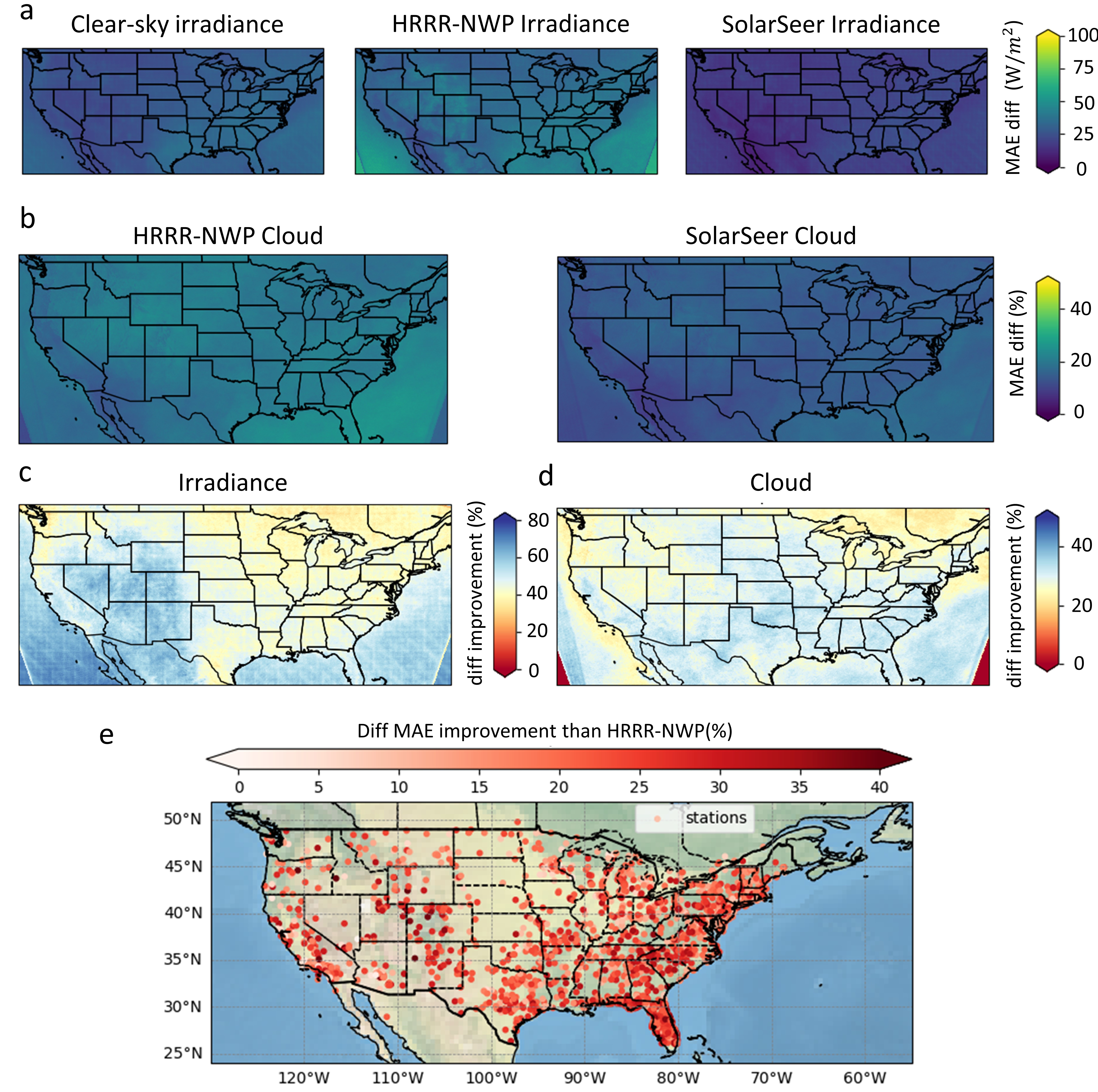}
\caption*{\justifying \textbf{Extended Data Fig. \thecustomfig \space \space} SolarSeer demonstrates smaller first-order difference forecasting MAE than HRRR-NWP. (a) MAE of the solar irradiance first-order difference forecasting. (b) MAE of the cloud cover first-order difference forecasting. (c) MAE improvement of the solar irradiance first-order difference forecasting. (d) MAE improvement of the cloud cover first-order difference forecasting. (e) MAE improvement of solar irradiance first-order difference forecasting in 1,800 weather station. All the initial time is UTC 06:00:00 because it approximately corresponds to local time 00:00:00.}
\label{extend_fig4}
\stepcounter{customfig}
\end{figure}

\begin{figure}[ht]
\centering
\includegraphics[width=\linewidth]{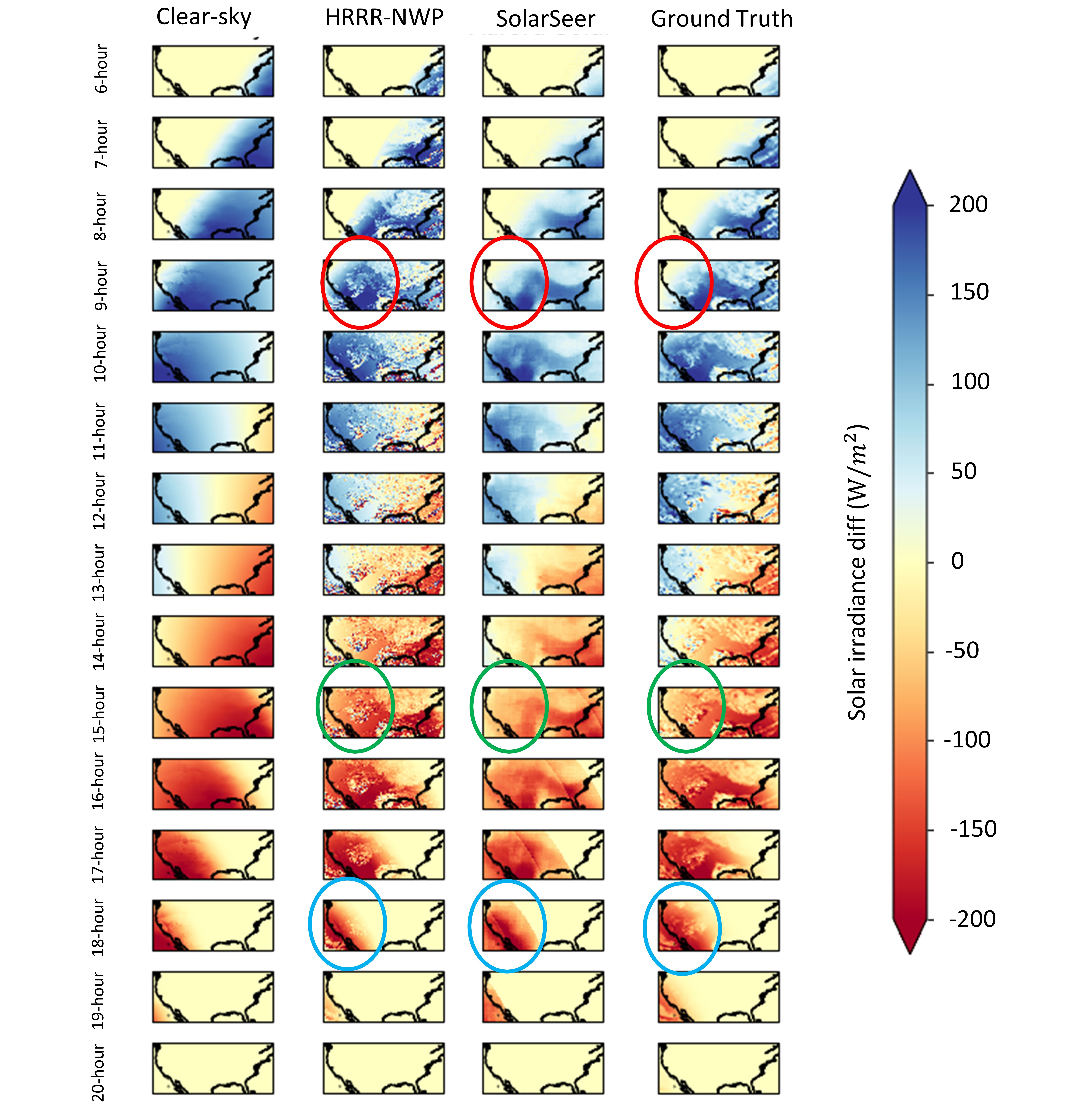}
\caption*{\justifying \textbf{Extended Data Fig. \thecustomfig \space \space} Visualization of solar irradiance first-order difference forecasting across the Contiguous United States. The initial time is UTC 2023-01-20 06:00:00. SolarSeer is closer to the ground truth than HRRR-NWP, particularly at lead times of 9, 15, and 18 hours.}
\label{extend_fig5}
\stepcounter{customfig}
\end{figure}

\begin{figure}[ht]
\centering
\includegraphics[width=\linewidth]{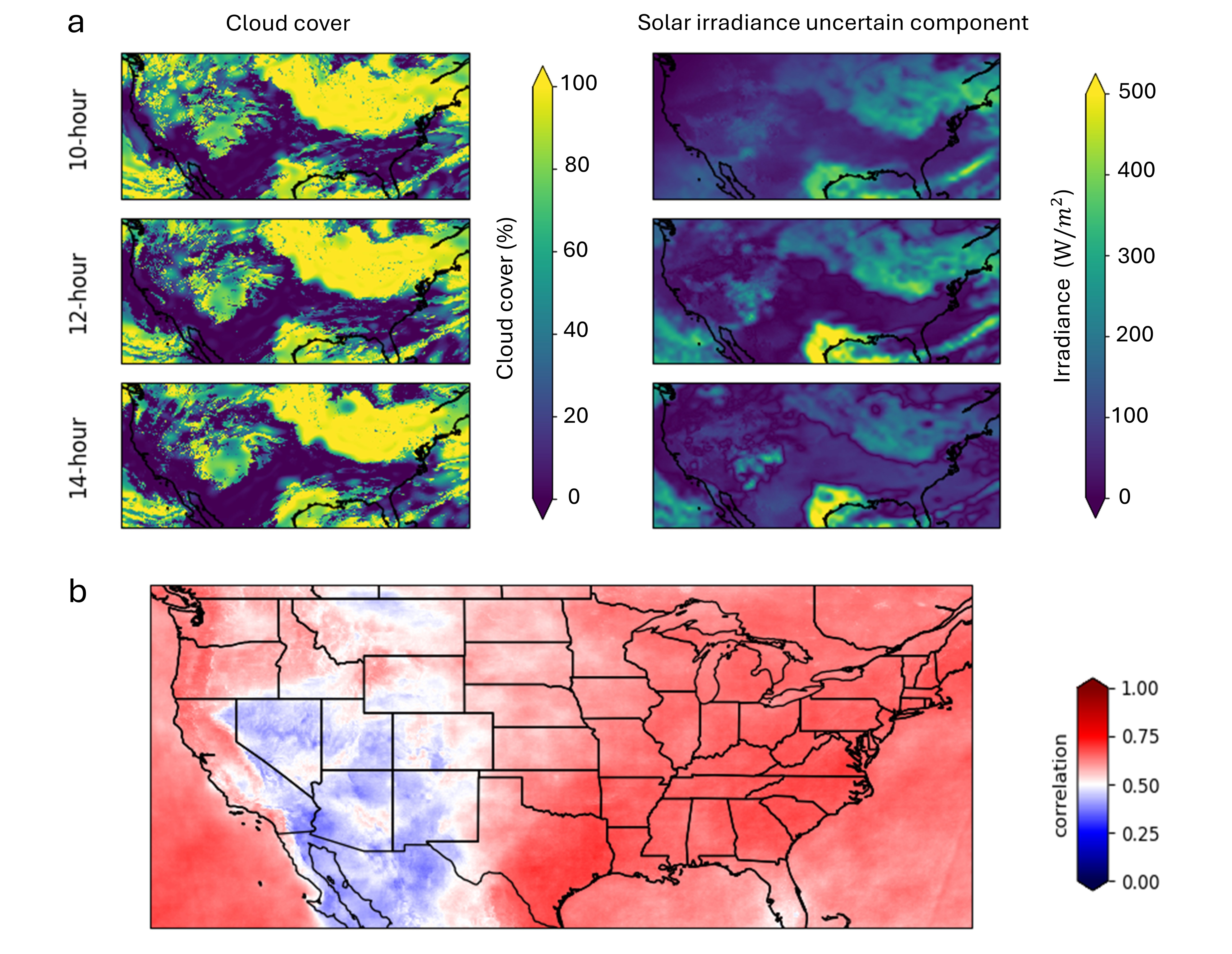}
\caption*{\justifying \textbf{Extended Data Fig. \thecustomfig \space \space} Interpretation of SolarSeer. (a) Visualization of cloud cover ground truth and solar irradiance uncertain component. (b) Pearson correlation coefficient between cloud cover ground truth and solar irradiance uncertain component during the day. The initial time is UTC 2023-01-20 06:00:00.}
\label{extend_fig6}
\stepcounter{customfig}
\end{figure}

As shown in \hyperref[fig5]{Fig.5} and \hyperref[extend_fig4]{Extended Data Fig. 4}, SolarSeer outperforms HRRR-NWP by reducing the MAE and RMSE of first-order irradiance difference forecasts in 100\% of regions. Furthermore, SolarSeer reduces the MAE of first-order cloud cover difference forecasts in 99.13\% of the regions and the RMSE in 99.82\% of the regions. On average, SolarSeer decreases the MAE and RMSE of first-order irradiance differences by 48.99\% and 50.12\%, respectively, in reanalysis data. It also reduces the MAE and RMSE of first-order cloud cover differences by 28.84\% and 28.93\%, respectively, in reanalysis data. \hyperref[extend_fig5]{Extended Data Fig. 5} compares the solar irradiance first-order difference forecasting from SolarSeer, HRRR-NWP, and clear-sky irradiance models against the ground truth across lead times of 6 to 20 hours. SolarSeer demonstrates closer agreement with the ground truth, particularly at lead times of 9, 15, and 18 hours (circled in \hyperref[extend_fig5]{Extended Data Fig. 5}).

We also verified the first-order irradiance difference forecasting results on 1,800 stations provided by Synoptic (see \hyperref[data]{Data}). As shown in \hyperref[fig5]{Fig.5}(e), \hyperref[fig5]{Fig.5}(f), and \hyperref[extend_fig4]{Extended Data Fig. 4}(e), SolarSeer significantly reduces the MAE of the first-order irradiance difference at 99.94\% of stations and the RMSE at 100\% of stations, with mean reductions of 20.50\% in MAE and 16.90\% in RMSE relative to HRRR-NWP at the initial time of UTC 06:00. The improvement at the stations is smaller than that observed in the reanalysis data, primarily because SolarSeer is trained using the reanalysis data as labels, and there are distributional discrepancies between the reanalysis data and the observations of the stations. 

SolarSeer significantly improves the forecasting of solar irradiance $Diff$ in both reanalysis data and station observation data. Improved forecasting of solar irradiance fluctuations can provide more useful information for tackling the uncertainty of solar PV power generation and thus help enhance the economic operation of future high-penetration PV power systems~\cite{gandhi2024value}.

\section*{Discussion}
\phantomsection
\addcontentsline{toc}{section}{Discussion}
\label{Discussion}
\hspace*{1em}
We present SolarSeer, an end-to-end large AI model for 24-hour-ahead solar energy forecasting across the Contiguous United States (CONUS). SolarSeer generates the future 24-hour forecasting of cloud cover and solar irradiance hourly for the CONUS at 5-kilometer spatial resolution. It significantly outperforms the state-of-the-art HRRR-NWP across CONUS in both computational speed and accuracy of 24-hour solar irradiance forecasts. The major contributions of SolarSeer are threefold. 

Firstly, SolarSeer is an ultrafast large AI model without data assimilation for 24-hour solar irradiance forecasting across the CONUS. SolarSeer designs an end-to-end large AI model that directly maps the historical satellite observations to the future cloud cover and solar irradiance forecasts, avoiding the significant computational costs of both data assimilation and PDE solving in NWP (\hyperref[framework]{Fig. 1} (a)-(c)). In under 3 seconds, SolarSeer can generate 24-hour forecasts of cloud cover and solar irradiance at a 5-kilometer spatial resolution using only one graphics card. In contrast, the state-of-the-art NWP models require one to several hours to generate 24-hour forecasts with supercomputer clusters. Our model, SolarSeer, runs over 1,500 times faster than NWP.

Secondly, SolarSeer is an accurate large AI model that outperforms the state-of-the-art NWP for 24-hour solar irradiance forecasts across the CONUS. Compared with HRRR-NWP, SolarSeer significantly reduces the root mean squared error (RMSE) of solar irradiance forecasting by 27.28\% in reanalysis data. Our results demonstrate that AI-based solar irradiance forecasting can outperform NWP for up to 24 hours using only historical satellite observations with large-scale spatial-temporal context, without relying on historical irradiance station observations or gridded initial fields generated by data assimilation as input. 

Thirdly, SolarSeer also effectively captures solar irradiance fluctuations and significantly enhances the forecasting accuracy of first-order irradiance difference. Compared with state-of-the-art NWP in the USA, SolarSeer decreases the RMSE of first-order irradiance difference forecasting by 50.12\% in reanalysis data. Improved forecasting of solar irradiance fluctuations can help the electricity grid tackle the uncertainty of solar PV power better. 

SolarSeer can substantially reduce computational costs and enhance the accuracy of 24-hour-ahead solar irradiance forecasts compared to state-of-the-art NWP models across the CONUS. The enhanced accuracy in solar irradiance forecasting can reduce the economic costs associated with forecasting errors, facilitate the optimal and economically efficient operation of future high-penetration solar PV power systems, and thereby promote the development of solar energy.

Future work will further improve the forecasting accuracy and expand SolarSeer to cover larger spatial and temporal forecasting horizons. Spatially, SolarSeer will be extended beyond the CONUS to other regions, with the ultimate goal of global solar irradiance forecasting. Temporally, the forecasting horizon will increase from 24 hours to 48 hours and eventually to several days, further advancing SolarSeer's contribution to high-penetration solar PV power systems.

\section*{Methods}
\phantomsection
\addcontentsline{toc}{section}{Methods}
\label{Methods}

\subsection*{Data}
\phantomsection
\addcontentsline{toc}{subsection}{Data}
\label{data}
\hspace*{1em}
SolarSeer leverages historical 6-hour satellite observations to generate 24-hour forecasts for total cloud cover and solar irradiance. The model is trained using satellite data as inputs, with RTMA reanalysis data serving as the cloud cover labels and ERA5 reanalysis data as the solar irradiance labels.

Satellite image data are obtained from the Geostationary Operational Environmental Satellite (GOES)~\cite{goes_noaa}, with a temporal resolution of 10 minutes. The Advanced Baseline Imager (ABI) on the GOES satellite captures visible and infrared images of the Earth. The ABI generates images across 16 different spectral bands. In this study, we used 4 specific bands, namely 2, 7, 10, and 14 bands (corresponding central shortwaves are 0.64, 3.9, 7.3, and 11.2 $\upmu$m) because they are highly related to clouds. The spatial resolution is 2 kilometers for channels 7, 10, and 14, and 0.5 kilometers for channel 2. 

The Real-Time Mesoscale Analysis (RTMA) dataset~\cite{de2011real} provides high-resolution analysis of near-surface weather conditions with a 2.5-kilometer resolution and hourly updates for the contiguous United States. We used the total cloud cover from the RTMA data as the ground truth of the cloud cover during network training and evaluation. The total cloud cover is between 0\% (clear sky) and 100\% (completely overcast). 

The fifth generation of ECMWF Reanalysis Data (ERA5)~\cite{hersbach2020era5} offers global gridded estimates at different atmospheric levels and near the surface with a spatial resolution of 0.25° and a temporal resolution of one hour. In the experiments, we used 'surface solar radiation downwards' from the ERA5 dataset as the ground truth for solar irradiance. 

We compared SolarSeer with state-of-the-art solar irradiance forecasting in the contiguous United States. The High-Resolution Rapid Refresh (HRRR)~\cite{dowell2022high, james2022high} model, developed by the National Oceanic and Atmospheric Administration (NOAA), is an hourly updated numerical weather prediction model with 3-kilometer resolution, designed for situational awareness and short-term forecasting. 

Our experiments used a spatial grid of 480 × 1150, encompassing latitudes from 26.00°N to 49.95°N and longitudes from 68.55°W to 126.00°W, with a spatial resolution of 0.05°. This grid covers most parts of the contiguous United States. GOES satellite, RTMA, ERA5, and HRRR-NWP data were standardized to a spatial resolution of 0.05 ° (approximately 5 km) and a temporal resolution of 1 hour by interpolation. 

Observation data from weather stations are usually closer to the actual solar irradiance than reanalysis data. Thus, we also evaluated SolarSeer in 1,800 weather stations across the Contiguous United States. Station observation data are provided by Synoptic Data, which brings together data from multiple sources worldwide, providing a comprehensive and accessible hub for critical environmental information. The data are accessed through their Weather API. For more details, please refer to Synoptic Data’s official site at \url{https://synopticdata.com/}. 

\subsection*{Clear-sky solar irradiance model}
\phantomsection
\addcontentsline{toc}{subsection}{Clear-sky solar irradiance model}
\label{clear-sky}
\hspace*{1em}
The total solar radiation on a horizontal surface is called global horizontal irradiance (GHI), which directly determines the PV power generation. Clear-sky GHI is used to model the maximum possible solar irradiance reaching the ground under cloud-free conditions~\cite{stein2012global}. This can be estimated using clear-sky models, among which the Ineichen-Perez model~\cite{ineichen2002new} is one of the most widely used models, largely due to its straightforward implementation via the Pvlib~\cite{holmgren2018pvlib} Python package. The Ineichen-Perez model is defined as follows~\cite{stein2012global}:
\begin{equation}
\begin{aligned}
{I_{clear}} &= {{\rm{c}}_{{\rm{g}}1}}{I_0}\cos (z)\exp \left\{ { - {{\rm{c}}_{{\rm{g}}2}}{\rm{AM}}\left[ {{{\rm{f}}_{{\rm{h}}1}} + {{\rm{f}}_{{\rm{h}}2}}({\rm{TL}} - 1)} \right]} \right\}\exp \left( {0.01{\rm{A}}{{\rm{M}}^{1.8}}} \right)\label{1}\\
{I_0} &= 1367.7 \times \left( {1 + 0.033 \times \cos \left( {\frac{{2\pi}}{{365}} \times \text{DOY}} \right)} \right)\\
{{\rm{c}}_{{\rm{g}}1}} &= 5.09 \times {10^{ - 5}} \times h + 0.868\\
{{\rm{c}}_{{\rm{g}}2}} &= 3.92 \times {10^{ - 5}} \times h + 0.0387\\
{{\rm{f}}_{{\rm{h}}1}} &= \exp ( - h/8000)\\
{{\rm{f}}_{{\rm{h}}2}} &= \exp ( - h/1250)\\
\text{AM} &= 1/\cos (z)
\end{aligned}
\end{equation}
where $I_{\text{clear}}$ is the clear-sky GHI, $\text{AM}$ is the air mass, $h$ is the elevation, $z$ is the zenith angle (a function of longitude, latitude, and time), $\text{DOY}$ is the day of the year, $I_0$ is the extraterrestrial radiation, and $\text{TL}$ is the Linke Turbidity (also a function of longitude, latitude, and time in the Pvlib library).

For a given place, the elevation h is generally fixed. Consequently, clear-sky GHI estimated by the Ineichen-Perez model depends on longitude, latitude, and time. Given the longitude and latitude of a location, the Ineichen-Perez model can be implemented to calculate the clear-sky Global Horizontal Irradiance (GHI) for any specific time, historical or future, such as “UTC 2025-12-31 12:00:00”, using a straightforward function call within the Python package Pvlib. Thus, the clear-sky GHI estimated by the Ineichen-Perez model can serve as valuable future prior knowledge for solar irradiance forecasting~\cite{yang2020choice}. Integrating cloud cover forecasts with clear-sky GHI estimates from the Ineichen-Perez model allows for highly accurate predictions of future solar irradiance. %By integrating cloud cover forecasts with clear-sky GHI estimates from the Ineichen-Perez model, future solar irradiance can be predicted with high accuracy. %

\subsection*{SolarSeer network framework }
\phantomsection
\addcontentsline{toc}{subsection}{SolarSeer network framework}
\label{network}
\hspace*{1em}
SolarSeer uses the historical 6-hour satellite observation data as inputs to generate the future 24-hour cloud cover and solar irradiance forecasting with a high spatial resolution of 0.05° (approximately 5 km), covering the geographical region bounded by 26.00°N, 126.00°W, 49.95°N, and 68.55°W (almost all parts of the contiguous United States). SolarSeer contains two blocks: the cloud block and the irradiance block. Detailed information of the SolarSeer framework is as follows (see \hyperref[framework]{Fig. 1} (d)):

The cloud block contains 4 Adaptive Fourier Neural Operator (AFNO)-like Transformer layers. Each layer integrates layer normalization, fast Fourier transformer (FFT), residual connections, multi-head attention, and inverse FFT. The reason for selecting AFNO~\cite{pathak2022fourcastnet} is that it can model long-range temporal dependencies and challenging partial differential equations well. The input data are embedded into high-dimensional feature representations through patch embedding, where the patch size is 4x4 and the embedding dimension is 600. The output of the last AFNO Transformer layer is up-sampled with patch recovery to restore the original resolution. To ensure cloud cover forecasts within the range of 0\% to 100\%, a Tanh activation function is applied at the output layer of SolarSeer’s cloud block.  

The irradiance block contains 8 Swin Transformer ~\cite{liu2022swin} layers. Each block integrates residual connections, layer normalization, and multi-head attention, where we used a patch size of 8x8, a window size of 16, an embedding dimension of 256, and 2 attention heads for each transformer layer. The input of the irradiance block is first embedded into high-dimensional feature representations through patch embedding and then up-sampled with patch recovery to restore the original resolution in the last layer of the irradiance block. 

SolarSeer has about 23 million parameters in total. Data span from May 13, 2017, to December 31, 2023. For training, we used data from May 13, 2017, to December 31, 2022, excluding the 4th-6th days of each month from January to October 2022, which were reserved for validation to adjust the hyperparameters of neural networks. The test set comprised the entire year of 2023 for comprehensive performance evaluation. SolarSeer is trained on Microsoft’s Azure AI platform with 16 AMD MI200 GPUs (1024GB graphical memory in total) for approximately one week. Through training in over five years of data, SolarSeer significantly reduces the average root mean squared error (RMSE) by 29.13\% for cloud cover forecasting and 27.28\% for solar irradiance forecasting in reanalysis data compared to the state-of-the-art HRRR-NWP model in the CONUS.

\subsection*{Evaluation metrics}
\phantomsection
\addcontentsline{toc}{subsection}{Evaluation metrics}
\label{metrics}
\hspace*{1em}
We evaluate the performance of solar irradiance forecasting and cloud cover forecasting using root mean squared error (RMSE) and mean absolute error (MAE) as follows:
\begin{equation}
RMSE =\sqrt{\frac{1}{n}\sum_{i=1}^{n}{(y_i-\widehat{y_i})}^2},\quad MAE =\frac{1}{n}\sum_{i=1}^{n}{|y_i-\widehat{y_i}|}\label{2}
\end{equation}
where $n$ is the number of data points in the test sets, $y_i$ and $\widehat{y_i}$ are the ground truth and predicted values, respectively. MAE and RMSE measure the differences between the prediction and the ground truth. Smaller MAE and RMSE mean better forecasting performance.

We also define the first-order difference metric $Diff$ to characterize the fluctuation of solar irradiance and cloud cover. 
\begin{equation}
Diff= y_t - y_{t-1}\label{10}
\end{equation}
where $y_t$ and $y_{t-1}$ are the solar irradiance or total cloud cover at the current time step and the previous time, respectively.

Based on the predicted values of solar irradiance and cloud cover, we can compute the predicted first-order difference of solar irradiance and cloud cover. The prediction performance of these first-order differences can then be evaluated using MAE and RMSE. 

\section*{Data availability}
\phantomsection
\addcontentsline{toc}{section}{Data availability}
\label{Data_availability}
\hspace*{1em}
We downloaded the ERA5 hourly solar irradiance data at the official website of the Copernicus Climate Data Store (CDS): \url{https://cds.climate.copernicus.eu}.
RTMA hourly total cloud cover data are available at: 
\url{https://emc.ncep.noaa.gov/emc/pages/numerical_forecast_systems/rtma.php}.
GOES satellite data are available at: \url{https://www.star.nesdis.noaa.gov/GOES/index.php}.
The historical HRRR forecast data are accessible via the Herbie Python package: \url{https://github.com/blaylockbk/Herbie}.
The real-time HRRR forecast data is available at:  
\url{https://nomads.ncep.noaa.gov/pub/data/nccf/com/hrrr/prod/}.

\section*{Code availability}
\phantomsection
\addcontentsline{toc}{section}{Code availability}
\label{Code_availability}
\hspace*{1em}
The code is available at \url{https://github.com/microsoft/SolarSeer/tree/main} and will be public once accepted. 
\section*{Acknowledgements}
\hspace*{1em}
The authors would like to thank ECMWF for providing publicly available ERA5 reanalysis data and NOAA for providing RTMA reanalysis data and GOES satellite data. We would also thank Synoptic Data PBC (linked to \url{https://synopticdata.com/}) for aggregating real-world weather station observations, which were essential for verifying the performance of our solar irradiance forecasts. 

\section*{Author contributions} 
\phantomsection
\addcontentsline{toc}{section}{Author contributions}
\label{Author_contributions}
\hspace*{1em}
M.B. trained the AI model, plotted figures, and wrote the manuscript. Z.F. processed the satellite data and trained the AI model. S.T. reviewed and polished the manuscript and figures. S. X., P.Z., and  W. J. processed the reanalysis data and the station observation data as well as improved the AI model. J.B., Y.X., J.W., B.Z., H.S., K.T., and Q.Z. supported model training and reviewed the manuscript. H.D., X.Z., H.S., and Q.W. reviewed, discussed, and supervised this work. 

\section*{Competing interests}
\phantomsection
\addcontentsline{toc}{section}{Competing interests}
\label{Competing_interests}
\hspace*{1em}
The authors declare no competing interests. 

\bibliography{ref}

\begin{thebibliography}{10}
\urlstyle{rm}
\expandafter\ifx\csname url\endcsname\relax
  \def\url#1{\texttt{#1}}\fi
\expandafter\ifx\csname urlprefix\endcsname\relax\def\urlprefix{URL }\fi
\expandafter\ifx\csname doiprefix\endcsname\relax\def\doiprefix{DOI: }\fi
\providecommand{\bibinfo}[2]{#2}
\providecommand{\eprint}[2][]{\url{#2}}

\bibitem{IEA2023}
\bibinfo{author}{{International Energy Agency}}.
\newblock \bibinfo{title}{{World Energy Outlook 2023}}.
\newblock \bibinfo{type}{Tech. Rep.}, \bibinfo{institution}{{IEA}} (\bibinfo{year}{2023}).

\bibitem{zhang2025machine}
\bibinfo{author}{Zhang, Z.} \emph{et~al.}
\newblock \bibinfo{journal}{\bibinfo{title}{A machine learning model for hub-height short-term wind speed prediction}}.
\newblock {\emph{\JournalTitle{Nature Communications}}} \textbf{\bibinfo{volume}{16}}, \bibinfo{pages}{3195} (\bibinfo{year}{2025}).

\bibitem{solarpowereurope2024}
\bibinfo{author}{{SolarPower Europe}}.
\newblock \bibinfo{title}{{Global Market Outlook for Solar Power 2024-2028}}.
\newblock \bibinfo{type}{Tech. Rep.}, \bibinfo{institution}{{SolarPower Europe}} (\bibinfo{year}{2024}).

\bibitem{nijsse2023momentum}
\bibinfo{author}{Nijsse, F.~J.} \emph{et~al.}
\newblock \bibinfo{journal}{\bibinfo{title}{The momentum of the solar energy transition}}.
\newblock {\emph{\JournalTitle{Nature Communications}}} \textbf{\bibinfo{volume}{14}}, \bibinfo{pages}{6542} (\bibinfo{year}{2023}).

\bibitem{wang2023inherent}
\bibinfo{author}{Wang, J.} \emph{et~al.}
\newblock \bibinfo{journal}{\bibinfo{title}{Inherent spatiotemporal uncertainty of renewable power in china}}.
\newblock {\emph{\JournalTitle{Nature Communications}}} \textbf{\bibinfo{volume}{14}}, \bibinfo{pages}{5379} (\bibinfo{year}{2023}).

\bibitem{xia2024accurate}
\bibinfo{author}{Xia, P.} \emph{et~al.}
\newblock \bibinfo{journal}{\bibinfo{title}{Accurate nowcasting of cloud cover at solar photovoltaic plants using geostationary satellite images}}.
\newblock {\emph{\JournalTitle{Nature Communications}}} \textbf{\bibinfo{volume}{15}}, \bibinfo{pages}{510} (\bibinfo{year}{2024}).

\bibitem{gandhi2024value}
\bibinfo{author}{Gandhi, O.} \emph{et~al.}
\newblock \bibinfo{journal}{\bibinfo{title}{The value of solar forecasts and the cost of their errors: A review}}.
\newblock {\emph{\JournalTitle{Renewable and Sustainable Energy Reviews}}} \textbf{\bibinfo{volume}{189}}, \bibinfo{pages}{113915} (\bibinfo{year}{2024}).

\bibitem{bauer2015quiet}
\bibinfo{author}{Bauer, P.}, \bibinfo{author}{Thorpe, A.} \& \bibinfo{author}{Brunet, G.}
\newblock \bibinfo{journal}{\bibinfo{title}{The quiet revolution of numerical weather prediction}}.
\newblock {\emph{\JournalTitle{Nature}}} \textbf{\bibinfo{volume}{525}}, \bibinfo{pages}{47--55} (\bibinfo{year}{2015}).

\bibitem{zhou2024progress}
\bibinfo{author}{Zhou, W.} \emph{et~al.}
\newblock \bibinfo{journal}{\bibinfo{title}{Progress and future prospects of decadal prediction and data assimilation: a review}}.
\newblock {\emph{\JournalTitle{Atmospheric and Oceanic Science Letters}}} \textbf{\bibinfo{volume}{17}}, \bibinfo{pages}{100441} (\bibinfo{year}{2024}).

\bibitem{pu2019numerical}
\bibinfo{author}{Pu, Z.} \& \bibinfo{author}{Kalnay, E.}
\newblock \bibinfo{journal}{\bibinfo{title}{Numerical weather prediction basics: Models, numerical methods, and data assimilation}}.
\newblock {\emph{\JournalTitle{Handbook of hydrometeorological ensemble forecasting}}} \bibinfo{pages}{67--97} (\bibinfo{year}{2019}).

\bibitem{chen2024one}
\bibinfo{author}{Chen, Y.}, \bibinfo{author}{Lin, C.}, \bibinfo{author}{Liu, J.} \& \bibinfo{author}{Yu, D.}
\newblock \bibinfo{journal}{\bibinfo{title}{One-hour-ahead solar irradiance forecast based on real-time k-means++ clustering on the input side and cnn-lstm}}.
\newblock {\emph{\JournalTitle{Journal of Atmospheric and Solar-Terrestrial Physics}}} \bibinfo{pages}{106405} (\bibinfo{year}{2024}).

\bibitem{kumari2021deep}
\bibinfo{author}{Kumari, P.} \& \bibinfo{author}{Toshniwal, D.}
\newblock \bibinfo{journal}{\bibinfo{title}{Deep learning models for solar irradiance forecasting: A comprehensive review}}.
\newblock {\emph{\JournalTitle{Journal of Cleaner Production}}} \textbf{\bibinfo{volume}{318}}, \bibinfo{pages}{128566} (\bibinfo{year}{2021}).

\bibitem{zhu2025novel}
\bibinfo{author}{Zhu, L.}, \bibinfo{author}{Huang, X.}, \bibinfo{author}{Zhang, Z.}, \bibinfo{author}{Li, C.} \& \bibinfo{author}{Tai, Y.}
\newblock \bibinfo{journal}{\bibinfo{title}{A novel u-lstm-aft model for hourly solar irradiance forecasting}}.
\newblock {\emph{\JournalTitle{Renewable Energy}}} \textbf{\bibinfo{volume}{238}}, \bibinfo{pages}{121955} (\bibinfo{year}{2025}).

\bibitem{yang2022review}
\bibinfo{author}{Yang, D.} \emph{et~al.}
\newblock \bibinfo{journal}{\bibinfo{title}{A review of solar forecasting, its dependence on atmospheric sciences and implications for grid integration: Towards carbon neutrality}}.
\newblock {\emph{\JournalTitle{Renewable and Sustainable Energy Reviews}}} \textbf{\bibinfo{volume}{161}}, \bibinfo{pages}{112348} (\bibinfo{year}{2022}).

\bibitem{srivastava2011electricity}
\bibinfo{author}{Srivastava, A.~K.}, \bibinfo{author}{Kamalasadan, S.}, \bibinfo{author}{Patel, D.}, \bibinfo{author}{Sankar, S.} \& \bibinfo{author}{Al-Olimat, K.~S.}
\newblock \bibinfo{journal}{\bibinfo{title}{Electricity markets: an overview and comparative study}}.
\newblock {\emph{\JournalTitle{International Journal of Energy Sector Management}}} \textbf{\bibinfo{volume}{5}}, \bibinfo{pages}{169--200} (\bibinfo{year}{2011}).

\bibitem{bi2023accurate}
\bibinfo{author}{Bi, K.} \emph{et~al.}
\newblock \bibinfo{journal}{\bibinfo{title}{Accurate medium-range global weather forecasting with 3d neural networks}}.
\newblock {\emph{\JournalTitle{Nature}}} \textbf{\bibinfo{volume}{619}}, \bibinfo{pages}{533--538} (\bibinfo{year}{2023}).

\bibitem{lam2023learning}
\bibinfo{author}{Lam, R.} \emph{et~al.}
\newblock \bibinfo{journal}{\bibinfo{title}{Learning skillful medium-range global weather forecasting}}.
\newblock {\emph{\JournalTitle{Science}}} \textbf{\bibinfo{volume}{382}}, \bibinfo{pages}{1416--1421} (\bibinfo{year}{2023}).

\bibitem{price2025probabilistic}
\bibinfo{author}{Price, I.} \emph{et~al.}
\newblock \bibinfo{journal}{\bibinfo{title}{Probabilistic weather forecasting with machine learning}}.
\newblock {\emph{\JournalTitle{Nature}}} \textbf{\bibinfo{volume}{637}}, \bibinfo{pages}{84--90} (\bibinfo{year}{2025}).

\bibitem{chen2023fuxi}
\bibinfo{author}{Chen, L.} \emph{et~al.}
\newblock \bibinfo{journal}{\bibinfo{title}{Fuxi: A cascade machine learning forecasting system for 15-day global weather forecast}}.
\newblock {\emph{\JournalTitle{npj Climate and Atmospheric Science}}} \textbf{\bibinfo{volume}{6}}, \bibinfo{pages}{190} (\bibinfo{year}{2023}).

\bibitem{bodnar2025foundation}
\bibinfo{author}{Bodnar, C.} \emph{et~al.}
\newblock \bibinfo{journal}{\bibinfo{title}{A foundation model for the earth system}}.
\newblock {\emph{\JournalTitle{Nature}}} \bibinfo{pages}{1--8} (\bibinfo{year}{2025}).

\bibitem{espeholt2022deep}
\bibinfo{author}{Espeholt, L.} \emph{et~al.}
\newblock \bibinfo{journal}{\bibinfo{title}{Deep learning for twelve hour precipitation forecasts}}.
\newblock {\emph{\JournalTitle{Nature communications}}} \textbf{\bibinfo{volume}{13}}, \bibinfo{pages}{1--10} (\bibinfo{year}{2022}).

\bibitem{fisher2009data}
\bibinfo{author}{Fisher, M.}, \bibinfo{author}{Nocedal, J.}, \bibinfo{author}{Tr{\'e}molet, Y.} \& \bibinfo{author}{Wright, S.~J.}
\newblock \bibinfo{journal}{\bibinfo{title}{Data assimilation in weather forecasting: a case study in pde-constrained optimization}}.
\newblock {\emph{\JournalTitle{Optimization and Engineering}}} \textbf{\bibinfo{volume}{10}}, \bibinfo{pages}{409--426} (\bibinfo{year}{2009}).

\bibitem{schultz2021can}
\bibinfo{author}{Schultz, M.~G.} \emph{et~al.}
\newblock \bibinfo{journal}{\bibinfo{title}{Can deep learning beat numerical weather prediction?}}
\newblock {\emph{\JournalTitle{Philosophical Transactions of the Royal Society A}}} \textbf{\bibinfo{volume}{379}}, \bibinfo{pages}{20200097} (\bibinfo{year}{2021}).

\bibitem{dowell2022high}
\bibinfo{author}{Dowell, D.~C.} \emph{et~al.}
\newblock \bibinfo{journal}{\bibinfo{title}{The high-resolution rapid refresh (hrrr): An hourly updating convection-allowing forecast model. part i: Motivation and system description}}.
\newblock {\emph{\JournalTitle{Weather and Forecasting}}} \textbf{\bibinfo{volume}{37}}, \bibinfo{pages}{1371--1395} (\bibinfo{year}{2022}).

\bibitem{james2022high}
\bibinfo{author}{James, E.~P.} \emph{et~al.}
\newblock \bibinfo{journal}{\bibinfo{title}{The high-resolution rapid refresh (hrrr): an hourly updating convection-allowing forecast model. part ii: Forecast performance}}.
\newblock {\emph{\JournalTitle{Weather and Forecasting}}} \textbf{\bibinfo{volume}{37}}, \bibinfo{pages}{1397--1417} (\bibinfo{year}{2022}).

\bibitem{de2011real}
\bibinfo{author}{De~Pondeca, M.~S.} \emph{et~al.}
\newblock \bibinfo{journal}{\bibinfo{title}{The real-time mesoscale analysis at noaa’s national centers for environmental prediction: current status and development}}.
\newblock {\emph{\JournalTitle{Weather and Forecasting}}} \textbf{\bibinfo{volume}{26}}, \bibinfo{pages}{593--612} (\bibinfo{year}{2011}).

\bibitem{marquis2015public}
\bibinfo{author}{Marquis, M.}, \bibinfo{author}{Benjamin, S.}, \bibinfo{author}{James, E.}, \bibinfo{author}{Molling, C.} \emph{et~al.}
\newblock \bibinfo{title}{A public-private-academic partnership to advance solar power forecasting}.
\newblock \bibinfo{type}{Tech. Rep.}, \bibinfo{institution}{NOAA ESRL} (\bibinfo{year}{2015}).

\bibitem{james2017unified}
\bibinfo{author}{James, E.~P.}, \bibinfo{author}{Benjamin, S.~G.} \& \bibinfo{author}{Marquis, M.}
\newblock \bibinfo{journal}{\bibinfo{title}{A unified high-resolution wind and solar dataset from a rapidly updating numerical weather prediction model}}.
\newblock {\emph{\JournalTitle{Renewable Energy}}} \textbf{\bibinfo{volume}{102}}, \bibinfo{pages}{390--405} (\bibinfo{year}{2017}).

\bibitem{zhang2022solar}
\bibinfo{author}{Zhang, G.}, \bibinfo{author}{Yang, D.}, \bibinfo{author}{Galanis, G.} \& \bibinfo{author}{Androulakis, E.}
\newblock \bibinfo{journal}{\bibinfo{title}{Solar forecasting with hourly updated numerical weather prediction}}.
\newblock {\emph{\JournalTitle{Renewable and Sustainable Energy Reviews}}} \textbf{\bibinfo{volume}{154}}, \bibinfo{pages}{111768} (\bibinfo{year}{2022}).

\bibitem{hersbach2020era5}
\bibinfo{author}{Hersbach, H.} \emph{et~al.}
\newblock \bibinfo{journal}{\bibinfo{title}{The era5 global reanalysis}}.
\newblock {\emph{\JournalTitle{Quarterly Journal of the Royal Meteorological Society}}} \textbf{\bibinfo{volume}{146}}, \bibinfo{pages}{1999--2049} (\bibinfo{year}{2020}).

\bibitem{kamadinata2020solar}
\bibinfo{author}{Kamadinata, J.~O.}, \bibinfo{author}{Ken, T.~L.} \& \bibinfo{author}{Suwa, T.}
\newblock \bibinfo{journal}{\bibinfo{title}{Solar irradiance fluctuation prediction methodology using artificial neural networks}}.
\newblock {\emph{\JournalTitle{Journal of Solar Energy Engineering}}} \textbf{\bibinfo{volume}{142}}, \bibinfo{pages}{031003} (\bibinfo{year}{2020}).

\bibitem{goes_noaa}
\bibinfo{author}{{National Oceanic and Atmospheric Administration}}.
\newblock \bibinfo{title}{{GOES - Geostationary Operational Environmental Satellite}}.
\newblock \bibinfo{howpublished}{{https://www.goes.noaa.gov/index.html}}.

\bibitem{stein2012global}
\bibinfo{author}{Stein, J.~S.}, \bibinfo{author}{Hansen, C.~W.} \& \bibinfo{author}{Reno, M.~J.}
\newblock \bibinfo{title}{Global horizontal irradiance clear sky models: implementation and analysis.}
\newblock \bibinfo{type}{Tech. Rep.}, \bibinfo{institution}{Sandia National Laboratories (SNL), Albuquerque, NM, and Livermore, CA~} (\bibinfo{year}{2012}).

\bibitem{ineichen2002new}
\bibinfo{author}{Ineichen, P.} \& \bibinfo{author}{Perez, R.}
\newblock \bibinfo{journal}{\bibinfo{title}{A new airmass independent formulation for the linke turbidity coefficient}}.
\newblock {\emph{\JournalTitle{Solar energy}}} \textbf{\bibinfo{volume}{73}}, \bibinfo{pages}{151--157} (\bibinfo{year}{2002}).

\bibitem{holmgren2018pvlib}
\bibinfo{author}{Holmgren, W.~F.}, \bibinfo{author}{Hansen, C.~W.} \& \bibinfo{author}{Mikofski, M.~A.}
\newblock \bibinfo{journal}{\bibinfo{title}{pvlib python: A python package for modeling solar energy systems}}.
\newblock {\emph{\JournalTitle{Journal of Open Source Software}}} \textbf{\bibinfo{volume}{3}}, \bibinfo{pages}{884} (\bibinfo{year}{2018}).

\bibitem{yang2020choice}
\bibinfo{author}{Yang, D.}
\newblock \bibinfo{journal}{\bibinfo{title}{Choice of clear-sky model in solar forecasting}}.
\newblock {\emph{\JournalTitle{Journal of Renewable and Sustainable Energy}}} \textbf{\bibinfo{volume}{12}} (\bibinfo{year}{2020}).

\bibitem{pathak2022fourcastnet}
\bibinfo{author}{Pathak, J.} \emph{et~al.}
\newblock \bibinfo{journal}{\bibinfo{title}{Fourcastnet: A global data-driven high-resolution weather model using adaptive fourier neural operators}}.
\newblock {\emph{\JournalTitle{arXiv preprint arXiv:2202.11214}}}  (\bibinfo{year}{2022}).

\bibitem{liu2022swin}
\bibinfo{author}{Liu, Z.} \emph{et~al.}
\newblock \bibinfo{title}{Swin transformer v2: Scaling up capacity and resolution}.
\newblock In \emph{\bibinfo{booktitle}{Proceedings of the IEEE/CVF conference on computer vision and pattern recognition}}, \bibinfo{pages}{12009--12019} (\bibinfo{year}{2022}).

\end{thebibliography}

\begin{figure}[ht]
\centering
\includegraphics[width=\linewidth]{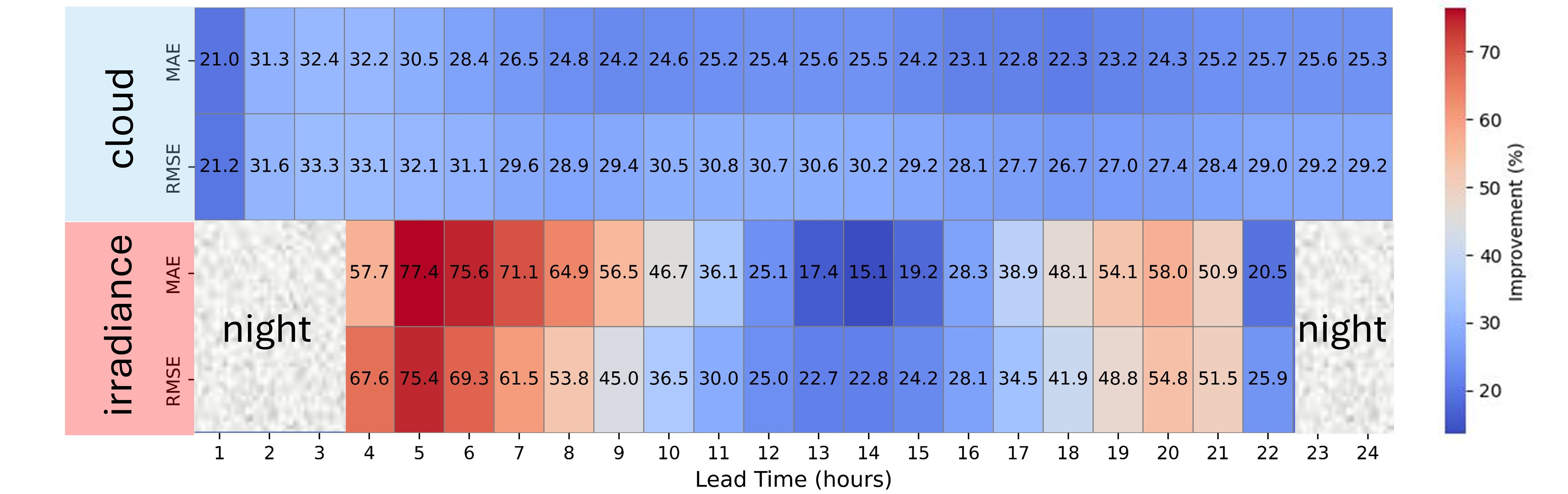}
\caption*{\textbf{Extended Data Fig. \thecustomfig \space \space} Improvement percentage of SolarSeer than HRRR-NWP. The initial time is UTC 06:00:00.}
\label{extend_fig7}
\stepcounter{customfig}
\end{figure}

\end{document}